\documentclass{article}

\PassOptionsToPackage{numbers,compress,sort}{natbib}
\usepackage[main, final]{neurips_2026}

\usepackage[utf8]{inputenc} 
\usepackage[T1]{fontenc}    
\usepackage{hyperref}       
\usepackage{url}            
\usepackage{booktabs}       
\usepackage{amsfonts}       
\usepackage{nicefrac}       
\usepackage{microtype}      
\usepackage{xcolor}         

\usepackage{microtype}
\usepackage{graphicx}
\usepackage{subcaption}
\usepackage{booktabs} 
\usepackage[most]{tcolorbox}
\tcbuselibrary{skins}

\usepackage{amsmath}
\usepackage{float}
\usepackage{amssymb}
\usepackage{mathtools}
\usepackage{amsthm}
\usepackage{xspace}
\usepackage{booktabs}
\usepackage[most]{tcolorbox}
\tcbuselibrary{skins}
\usepackage{cleveref}
\usepackage{wrapfig}
\usepackage{graphicx}
\usepackage{multirow}
\usepackage{amsfonts}
\usepackage{amsthm}
\usepackage{extarrows}
\usepackage{wrapfig}

\theoremstyle{plain}
\newtheorem{theorem}{Theorem}[section]
\newtheorem{proposition}[theorem]{Proposition}

\theoremstyle{definition}
\newtheorem{definition}[theorem]{Definition}

\theoremstyle{remark}

\definecolor{firstblue}{RGB}{0, 76, 153} 
\definecolor{secondgreen}{RGB}{238, 177, 86} 
\definecolor{thirdorange}{RGB}{219, 79, 64} 

\definecolor{firstblue}{RGB}{114, 63, 255} 
\definecolor{secondgreen}{RGB}{6, 7, 10} 
\definecolor{thirdorange}{RGB}{255, 192, 29} 

\definecolor{firstblue}{RGB}{63, 19, 8} 
\definecolor{secondgreen}{RGB}{191, 110, 63} 
\definecolor{thirdorange}{RGB}{216, 150, 90} 

\newcommand{\first}[1]{\textcolor{firstblue}{\textbf{#1}}}
\newcommand{\second}[1]{\textcolor{secondgreen}{\textbf{#1}}}
\newcommand{\third}[1]{\textcolor{thirdorange}{\textbf{#1}}}

\newcommand{\gnn}{\textsc{GNN}\xspace}
\newcommand{\our}{\textsc{SP4LP}\xspace}

\newcommand{\meanstd}[2]{#1 {\scriptsize ($\pm$ #2)}}

\newcommand{\fra}[1]{{\color{black}#1}}

\definecolor{softblue}{RGB}{47, 109, 166}
\definecolor{darkred}{RGB}{139, 0, 0}
\newcommand{\revision}[1]{\textcolor{black}{#1}}

\hypersetup{
    colorlinks=true,
    citecolor=softblue,
    linkcolor=darkred,
    urlcolor=softblue
}

\usepackage[textsize=tiny]{todonotes}

\title{GNNs Meet Sequence Models Along the Shortest-Path: \\ an Expressive Method for Link Prediction}

\author{%
  Francesco Ferrini\thanks{Equal contribution. Corresponding authors: \texttt{fferrini@fbk.eu}, \texttt{veronica.lachi@uit.no}}\hspace{4pt}\textsuperscript{,1} \quad
  Veronica Lachi\footnotemark[1]\hspace{4pt}\textsuperscript{,2} \quad
  Antonio Longa\textsuperscript{2} \quad
  Bruno Lepri\textsuperscript{1} \quad
  Andrea Passerini\textsuperscript{3} \\
  \vspace{0.2cm} \\
  \textsuperscript{1}Fondazione Bruno Kessler, Trento, Italy \\
  \textsuperscript{2}The Arctic University of Norway, Tromsø, Norway \\
  \textsuperscript{3}University of Trento, Trento, Italy
}

\begin{document}

\maketitle

\begin{abstract}
  Graph Neural Networks (GNNs) often fail to capture the link-specific structural patterns essential for accurate link prediction, since their node-centric message passing might overlook the subgraph structures connecting two nodes. Prior attempts to inject such structural context either suffer from high computational cost or rely on oversimplified heuristics (e.g., common neighbor counts) that cannot capture multi-hop dependencies. We propose \our (Shortest Path for Link Prediction), a new framework that integrates GNN-based node encodings with sequence modeling over shortest paths. Specifically, \our first computes node representations with a GNN, then extracts the shortest path between each candidate node pair and processes the sequence of node embeddings with a sequence model. This design allows \our to efficiently capture expressive multi-hop relational patterns. Theoretically, we show that \our is strictly more expressive than both standard message-passing GNNs and several leading structural feature methods, positioning it as a general and principled framework for link prediction in graphs. Empirically, \our sets a new state of the art on many standard link prediction benchmarks. 
\end{abstract}

\section{Introduction}
\label{sec:intro}
Graph Neural Networks (GNNs) are widely adopted for link-level tasks such as link prediction~\citep{zhang2018link,lu2011link,zhou2021progresses,NEURIPS2025_b1cd7227}, link classification~\citep{rossi2021knowledge,wang2021apan,cheng2025edge} and link regression~\citep{liang2025line,dong2019link} with applications spanning recommender systems~\citep{ying2018graph}, knowledge graph completion~\citep{nickel2015review}, and biological interaction prediction~\citep{jha2022prediction}. 

\begin{wrapfigure}{r}{0.30\textwidth}
    \centering
    \includegraphics[width=0.29\textwidth]{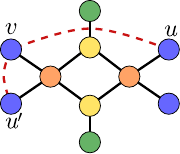}
    \caption{Links $(v,u)$ and $(v,u')$ have different structural roles within the graph, yet a GNN assigns them identical representations.}
    \label{fig:myfigure}
\end{wrapfigure}

Despite their popularity, standard GNNs struggle to accurately represent links, as they typically construct link embeddings by aggregating the representations of the two endpoint nodes. This node-centric strategy leads to a key limitation: structurally distinct links may be mapped to the same representation when their endpoints are automorphic \citep{srinivasan2019equivalence,chamberlaingraph,zhang2021labeling}. For example, in the graph of Figure \ref{fig:myfigure}, links $(v,u)$ and $(v,u')$ yield identical representations under any standard GNN, even if one pair shares a common neighbor and the other does not. This issue, known as the automorphic node problem~\citep{chamberlaingraph}, highlights a fundamental expressivity bottleneck in message-passing schemes for link representation.
 
To address this, several methods enhance GNNs with structural features (SFs), which can be broadly classified into three paradigms~\citep{wangneural}: SF-then-GNN, which injects structural context into the graph before message passing (e.g., SEAL~\citep{zhang2021labeling}, NBFNet~\citep{zhu2021neural}); SF-and-GNN, which computes SFs and node embeddings in parallel (e.g., Neo-GNN~\citep{yun2021neo}, BUDDY~\citep{chamberlaingraph}, MPLP~\citep{dong2024pure}); and GNN-then-SF, which applies message passing once to compute node representations and then combines them using task-specific structural context (e.g., NCN and NCNC~\citep{wangneural}). 

While SF-then-GNN methods are expressive, they are computationally inefficient, often requiring subgraph extraction or retraining per link. SF-and-GNN models are efficient but rely on predefined heuristics, limiting their ability to capture rich relational patterns. GNN-then-SF approaches offer a compelling trade-off between expressivity and scalability, but current methods in this class, i.e., NCN and NCNC, still rely on overlap between the neighborhoods of the endpoints: NCN uses common neighbors, while NCNC also leverages predicted second-order common neighbors. In the absence of such overlapping neighbors, they revert to standard GNN behavior and lose expressive power.

In this paper we propose \our, a novel method in the GNN-then-SF paradigm that combines high expressiveness with computational efficiency. \our constructs a path-aware representation by incorporating the embeddings of all nodes along the shortest path connecting the two endpoints. These node embeddings, obtained via a base GNN, are then processed as a sequence using a dedicated sequence model, such as a Transformer~\citep{vaswani2017attention}, LSTM~\citep{hochreiter1997long}, or an injective summation function~\citep{xu2019powerful}. 

The first key advantage of \our is its \textbf{expressiveness}. Unlike structural features such as common neighbors, which may not exist for many node pairs and can lead to degenerate cases in sparse graphs, the shortest path is always defined between any two nodes if the graph is connected. Shortest paths are more broadly defined, as common neighbors imply a path, but not vice versa.
Moreover, since the embeddings of the nodes along the path are generated through message passing, they implicitly encode the broader local structure surrounding the link. This richer structural context enables \our to distinguish non-automorphic links even when their endpoints are automorphic, thereby overcoming the automorphic node problem. We formally prove that \our is strictly more expressive than \fra{several} existing approaches.

\our is \textbf{efficient} and \textbf{scalable}. The message-passing step is performed only once on the entire graph, and the shortest path computation is a preprocessing step. Unlike SF-then-GNN methods such as SEAL or NBFNet, \our avoids costly per-link subgraph extraction or online traversal during inference, allowing it to scale to large graphs and high-throughput settings.

Moreover, \our is a \textbf{general and flexible framework}. It can be instantiated with any GNN architecture to compute node embeddings (e.g., GCN~\citep{kipf2016semi}, GAT~\citep{veličković2018graph}, GraphSAGE~\citep{hamilton2017inductive}), and supports a range of sequence models for encoding the path structure, from lightweight aggregators to fully expressive recurrent or attention-based models.

Finally, \our achieves \textbf{state-of-the-art performance} across several benchmark datasets. Under the challenging HeaRT evaluation protocol~\citep{li2023evaluating}, it consistently outperforms existing link prediction methods while maintaining competitive inference speed and low memory usage.

These properties make \our a principled and practical solution for learning expressive link representations in real-world graph learning scenarios.

\section{Preliminaries}
\begin{definition}[\textit{graph}]\label{graph}
A \textbf{graph} is a tuple $G=(V  ,E  ,\mathbf{X} ^0 )$ where $V  =\{1,\ldots,n\}$ is a set of nodes, $E  \subseteq V \times V $ is a set of edges and  $\mathbf{X} ^0 \in \mathbb{R}^{n\times f}$ is the node features matrix. To each graph is associated an adjacency matrix $\mathbf{A}  \in \{0,1\}^{n\times n}$ with $\mathbf{A}_{{i,j}}=1$ if and only if $(i,j)\in E  $. In this work, we consider simple, finite and undirected graphs.
\end{definition}

\begin{definition}[\textit{message passing}]\label{gnn}
Let  $G = (V , E , \mathbf{X}^0 )$ be a graph. In the \textbf{message passing} scheme, the representation of nodes $v\in V $ is iteratively updated as follows:  
\begin{align}
\mathbf{x}_v^0 = \mathbf{X}_{[v,:]}^0 \quad \mathbf{x}^l_v =\text{\small UPDATE}\left(\mathbf{x}_v^{l-1}, \text{\small AGGREGATE}\left(\{\mathbf{x}_u^{l-1} \mid u \in N(v)\}\right)\right)
\end{align}
  
where $N(v)$ is the first-order neighborhood of node $v$.
\end{definition}
Graph Neural Networks (GNNs) are a class of neural architectures that operate on graphs by iteratively updating node representations through the message passing scheme. It has been proven that GNNs are at most as effective as the Weisfeiler\textendash{}Lehman (WL) test in distinguishing between graphs~\citep{morris2019weisfeiler,xu2019powerful}.
\begin{definition}[GNN link representation model]\label{LRmodel}
A \textbf{GNN link representation model} $M$ is a class of functions 
\begin{equation}
F : ((u,v),G) \mapsto \mathbf{x}_{(u,v)}\in \mathbb{R}^d  
\end{equation}
which maps node pairs in $(u,v)\in V\times V$ to vector representations using the message passing scheme defined in Definition \ref{gnn}.
\end{definition}
Note that the pair \((u, v)\) belongs to \(V \times V\), meaning that we compute a representation for any node pair, what we refer to as a \textit{link}, regardless of whether an edge between them exists in \(E\). This general definition reflects the nature of the downstream tasks we aim to address once the link representation is available, most notably, link prediction, where the objective is to estimate the likelihood of a connection between arbitrary node pairs. To this end, the model learns representations for all possible pairs, not just those connected by an edge. A widely adopted approach for learning such representations is what we refer to as a \textit{pure GNN}, defined as follows:

\begin{definition}[\textit{pure GNN}]\label{pure}
 A \textbf{pure GNN} model calculates representation $\mathbf{x}_{(u,v)}\in \mathbb{R}^d$ for each pair of nodes $(u,v)$ with $u,v\in V $ as follows:
\begin{equation}\label{eq:standart NN_link_rep}
\mathbf{x}_{(u,v)}=g(\mathbf{x}_u^L, \mathbf{x}_v^L)
\end{equation}
where $g$ is an aggregation function and $\mathbf{x}_u^L, \mathbf{x}_v^L$ are the node representation of $u$ and $v$ learned by $L$ layers of message passing as defined in Definition \ref{gnn}.    
\end{definition}

Pure GNNs are inherently limited in terms of expressiveness. In particular even when the base GNN is the most powerful, they may assign the same representation to structurally different links. Consider, for example, the graph in Figure \ref{fig:myfigure}: the colors of the nodes indicate the colors produced by the WL algorithm; thus, using a most powerful GNN nodes $u,u'$ will be assigned to the same representation. As a result, no matter how expressive the aggregation function \(g\) is, the representations of the pairs \((v, u)\) and \((v, u')\) will be identical. However, the links \((v, u)\) and \((v, u')\) have different roles within the graph structure. We provide a formal definition of what it means for two links to be different.

\begin{definition}[\textit{node permutation}]\label{permutation}
A \textbf{node permutation} $\pi:\{1,\ldots,n\}\rightarrow \{1,\ldots,n\}$ is a bijective function that assigns a new index to each node of the graph. All the $n!$ possible node permutations constitute the permutation group $\Pi_n$. Given a subset of nodes $S\subseteq V  $, we define the permutation $\pi$ on $S$ as $\pi(S):=\{\pi(i)|i\in S\}$. Additionally, we define $\pi (\mathbf{A}  )$ as the matrix $\mathbf{A}  $ with rows and columns permutated based on $\pi$, i.e., $\pi(\mathbf{A}  )_{\pi(i), \pi(j)}=\mathbf{A}_{{i,j}}$.
\end{definition}

\begin{definition}[\textit{automorphism}]\label{automorphism}
An $\textbf{automorphism}$ on the graph
$G=(V,E,\mathbf{X}^0)$ is a permutation $\sigma\in\Pi_n$
such that $\sigma(\mathbf{A})=\mathbf{A}$
\fra{and $\mathbf{X}^0_{\sigma(i),:}=\mathbf{X}^0_{i,:}$
for every $i\in V$}.
All such \fra{feature-preserving} automorphisms constitute
the automorphism group $\Sigma_n^G$.
\fra{For constant node features, this reduces to the usual
structural automorphism group.}
\end{definition}

\begin{definition}[\textit{automorphic nodes}]\label{isomnode}
Let  $G = (V , E , \mathbf{X}^0 )$  be a graph and $\Sigma_n^G$ its automorphism group. Two nodes  $u, v\in V  $  are said to be \textbf{automorphic nodes} ( $u \simeq v$ ) if: 
\begin{equation}
   \exists \sigma \in \Sigma_n^G \quad \text{s.t.} \quad \sigma(\{u\}) = \{v\}.   
\end{equation}
\end{definition}

\begin{definition}[\textit{automorphic links}]\label{isomlink}
Let  $G = (V , E , \mathbf{X}^0 )$  be a graph and $\Sigma_n^G$ its automorphism group. Two pairs of nodes  $(u, v), (u', v') \in V  \times V $  are said to be \textbf{automorphic links} ( $(u, v) \simeq (u', v')$ ) if: 
\begin{equation}
   \exists \sigma \in \Sigma_n^G \quad \text{s.t.} \quad \sigma(\{u, v\}) = \{u', v'\}.   
\end{equation}

\end{definition}

\begin{proposition}\label{prop:auto}
Pure GNN methods suffer from the \textit{automorphic node problem}, i.e., for any graph $G=(V,E,\mathbf{X}^0)$, for pairs of links $(u,v),(u', v')\in V\times V $ such that there exist $\sigma_1\in \Sigma^G_n$ and $\sigma_2\in \Sigma^G_n$ with $\sigma_1(u)=u'$ and $\sigma_2(v)=v'$, $\mathbf{x}_{(u,v)}=\mathbf{x}_{(u',v')}$, independently of whether $(u,v)$ and $(u',v')$ are isomorphic, i.e, whether there exists $\sigma\in \Sigma^G_n$ with $\sigma(\{u, v\}) = \{u', v'\}$.
\end{proposition}

This limitation is well-known in the literature~\citep{chamberlaingraph,zhang2021labeling}. Importantly, it does not arise from the expressiveness bounds of GNNs, which are constrained by the WL test. Even considering higher-order GNNs, i.e., $k$-GNN~\citep{morris2019weisfeiler}, automorphic nodes will be assigned to the same representation as the $k$-WL algorithm preserves graph automorphisms for every $k$ \citep{lichter2025computational,dawar2020generalizations,cai1992optimal}. Thus, no standard GNN can distinguish non automorphic links composed by automorphic nodes.

To tackle this, several models have been proposed that enhance message passing by incorporating structural features~\citep{wangneural,zhu2021neural,chamberlaingraph,wangequivariant,zhang2021labeling}, thereby increasing the expressive power of the resulting link representations.
We provide a formal definition of what it means for one link representation model to be more expressive and strictly more expressive than another.

\begin{definition}[\textit{more expressive}]\label{moreex}
Let \( \mathcal{M}_1 \) and \( \mathcal{M}_2 \) be two link representation models (Def. \ref{LRmodel}). \( \mathcal{M}_2 \) is \textbf{more expressive} than \( \mathcal{M}_1 \) (\( \mathcal{M}_1 \preceq \mathcal{M}_2 \)) if, for any graph \( G = (V , E , \mathbf{X}^0 ) \) and any pair \( (u, v), (u', v') \in V  \times V  \) with \( (u, v) \not\simeq (u', v') \):  
\begin{equation}
\begin{aligned}
\exists F_1 \in \mathcal{M}_1 :
\; F_1((u, v), G) \neq F_1((u', v'), G)
\;\Rightarrow 
\exists F_2 \in \mathcal{M}_2 :
\; F_2((u, v), G) \neq F_2((u', v'), G).
\end{aligned}
\end{equation}

\end{definition}
\begin{definition}[\textit{strictly more expressive}]\label{strictmoreex}
Let \( \mathcal{M}_1 \) and \( \mathcal{M}_2 \) be two link representation models (Def.~\ref{LRmodel}). We say that \( \mathcal{M}_2 \) is \textbf{strictly more expressive} than \( \mathcal{M}_1 \) (\( \mathcal{M}_1 \prec \mathcal{M}_2 \)) if:
\begin{itemize}
    \item \( \mathcal{M}_2 \) is more expressive than \( \mathcal{M}_1 \) (Def.~\ref{moreex}), and
    \item there exists a graph \( G = (V, E, \mathbf{X}^0) \) and a pair of links \( (u, v), (u', v') \in V \times V \) with \( (u, v) \not\simeq (u', v') \) such that:
    \[
    \begin{aligned}
        \forall F_1 \in \mathcal{M}_1:& \quad F_1((u, v), G) = F_1((u', v'), G) \\
        \text{and} \quad \exists F_2 \in \mathcal{M}_2:& \quad F_2((u, v), G) \neq F_2((u', v'), G).
    \end{aligned}
    \]
\end{itemize}
\end{definition}

In section~\ref{sec:method} we introduce our model \our and demonstrate its improved expressive power in distinguishing structurally different links.

\begin{figure*}[t]
    \centering
    \includegraphics[width=\linewidth]{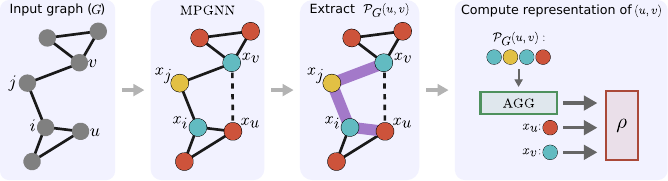}
    \caption{Overview of the \our framework. First, a GNN is used to compute contextualized embeddings for all nodes in the graph. Then, for each target link, the shortest path connecting the two endpoints is extracted. The embeddings of the nodes along this path are passed to a sequence model (e.g., Transformer or LSTM) to compute a path-aware link representation.}

    \label{fig:overview}
\end{figure*}

\section{Related Work}
\label{sec:related-work}
GNNs have been extensively used for link representation tasks. In standard approaches like Graph Autoencoders (GAE)~\citep{kipf2016variational}, node embeddings are computed via message passing, and a simple decoder (e.g., inner product followed by a sigmoid) predicts link existence. While efficient, these models exhibit limited expressiveness~\citep{zhang2021labeling,chamberlaingraph}, primarily due to their inability to capture rich structural patterns beyond immediate neighborhoods. This limitation has motivated the integration of explicit structural information into GNN-based models. \fra{An alternative approach~\citep{lachi2024a}, learns structural link representations by reformulating link prediction as node classification on a line graph.}

\textbf{Incorporating Structural Information into GNNs.}
Several methods inject structural information into GNN-based link predictors to overcome expressiveness bottlenecks.
Neo-GNN~\citep{yun2021neo} adds handcrafted structural signals to aggregation, while ELPH~\citep{chamberlaingraph} and its scalable variant \textsc{BUDDY} encode multi-hop neighborhood patterns via MinHash and HyperLogLog sketches, precomputed offline in \textsc{BUDDY} for efficiency.
\revision{MPLP~\citep{dong2024pure} replaces sketching with message passing over quasi-orthogonal random vectors, yielding unbiased estimates of common neighbor counts and Distance Encoding labels concatenated with GNN embeddings.}
NCN~\citep{wangneural} aggregates embeddings from endpoints and their common neighbors, and NCNC extends it by first predicting missing neighbors.
NBFNet~\citep{zhu2021neural} instead aggregates over all node-pair paths via Bellman--Ford-inspired recursive updates.
Beyond GNNs, PROXI~\citep{tola2025proxi} combines handcrafted proximity indices with XGBoost; while related through its use of structural signals, it fundamentally differs from GNN-based link representation models.
Despite these advances, such mechanisms still struggle to distinguish more complex structures, as shown in the next section.

\textbf{Enhancing GNN Expressiveness through Positional and Structural Encoding.}
Positional encodings further enrich GNN expressiveness. PEG~\citep{wangequivariant} integrates Laplacian-based encodings into message passing, weighting neighbors by positional distances. SEAL~\citep{zhang2021labeling} extracts $h$-hop enclosing subgraphs and labels nodes via the DRNL scheme before applying a GNN. While expressive, these methods struggle to scale due to subgraph extraction overhead.
In contrast, our model achieves expressiveness by operating directly on compact, informative shortest-path sequences, enabling better scalability without sacrificing representational power.


\textbf{Shortest-Path and Path-Based Methods in Graph Learning.}
Path-based representations have a long history in graph learning, from shortest-path kernels~\citep{borgwardt2005shortest} to recent neural approaches that aggregate information over walks or simple paths for graph- and node-level tasks~\citep{ying2021transformers,chen2020convolutional,michel2023path, chen2024learning}, and to generative graph transformers built on Eulerian path linearizations~\citep{zhao2025graphgpt}. Within link prediction, Geodesic GNN (GDGNN)~\citep{kong2022geodesic} aggregates node representations along geodesic connections after global message passing, typically using permutation-invariant aggregation that abstracts away ordering. In contrast, \our is pair-centric and explicitly models the shortest path between two endpoints as a learnable ordered sequence, yielding link representations that capture directional and positional structure. Recent results~\citep{airale2025simple} further show that walk-based encodings can fail to distinguish node pairs that path-based representations separate.

\section{\our: An Expressive GNN-then-SF Model for link Representation}\label{sec:method}

Existing GNN link representation models that leverage Structural Features (SF) fall into three categories~\citep{wangneural}: \textbf{SF-and-GNN}, which compute SF and GNN embeddings separately and then combine them (e.g., Neo-GNN~\citep{yun2021neo}, BUDDY~\citep{chamberlaingraph}); \textbf{SF-then-GNN}, which augment the graph with SF before applying GNN (e.g., SEAL~\citep{zhang2018link}, NBFNet~\citep{zhu2021neural}); and \textbf{GNN-then-SF}, which compute GNN embeddings first and then aggregate them using SF (e.g., NCN, NCNC~\citep{wangneural}).

In this work, we adopt the GNN-then-SF paradigm, which combines the scalability advantages of applying message passing only once, as in the SF-and-GNN setting, with the expressiveness typical of SF-then-GNN approaches, due to the ability to aggregate over task-specific sets of node representations. To date, the only existing models that follow this paradigm are NCN and NCNC. For a more in-depth discussion on the differences between our method, \our and NCN, refer to Section~\ref{sec:comparisonncn}.
In the following, we introduce the necessary definitions, formally describe the model, and present theoretical results characterizing its expressive power.

\begin{definition}[\textit{path}]
Let $G = (V , E , \mathbf{X}^0 )$ be a graph and $u,v\in V$ to nodes. A \textbf{path} in $G$ from $u$ to $v$ is a sequence of nodes $P= (u_0, u_1, \dots, u_k) $ with (i) $u_i\in V$ for all  $i = 0, \dots, k-1$, (ii) $ u_0 = u$ and $ u_k = v $, (iii) $ (u_i, u_{i+1}) \in E $ for all  $i = 0, \dots, k-1$ , and  (iv) all nodes in the sequence are distinct (i.e., $ u_i \neq u_j$ for all $i \neq j $).
The length of a path $P$, $\text{len}(P)$ is the number of edges it contains.
\end{definition}

\begin{definition}[\textit{shortest path length}]
Let $\mathcal{P}_G(u, v) $ denote the set of all paths from $ u$ to $ v$ in $ G$. The \textbf{shortest path length} $d_G(u, v) $ is the minimum length among all paths i.e., $d_G(u, v) = \min_{P \in \mathcal{P}(u, v)} \text{len}(P)$.

\end{definition}

\begin{definition}[\textit{shortest path}]
A \textbf{shortest path} between $ u$ to $ v$ in $ G$ is any path $P^* \in \mathcal{P}_G(u, v) $ such that $\text{len}(P^*) = d_G(u, v) $. The set of all the shortest path from $u$ to $v$ in $G$ is denoted as $\mathcal{P}^*_G(u, v)$.
\end{definition}

Let $G = (V , E , \mathbf{X}^0 )$ be a graph, $u,v\in V$. \our is a GNN link representation model (see Definition \ref{LRmodel}) that computes link representation as follows:


\begin{equation}\label{eq:our}
\resizebox{\textwidth}{!}{$
\our((u,v),G)= \rho\left(\gnn(u,G), \gnn(v,G), 
\text{AGG}\left(\left\{\fra{\phi\left((\gnn(u_i,G))_{i=0}^{k}\right)} \mid \fra{(u_i)_{i=0}^{k}} \in \mathcal{P}^*_G\{u, v\}\right\}\right)
\right)
$}
\end{equation}

where $k=d_G(u,v)$, $\gnn(u,G)\in \mathbb{R}^d$ is the representation of node $u\in V$ obtained at the final layer of message passing as in Definition \ref{gnn}, \fra{$\phi:\bigcup_{\ell\geq 1}(\mathbb{R}^d)^\ell \rightarrow \mathbb{R}^d$ is a sequence encoder operating on the full ordered sequence of node embeddings along the shortest path from $u$ to $v$, including both endpoints}, $\text{AGG}$ is an aggregation function over multiset of shortest paths representations and $\rho:\mathbb{R}^d\times\mathbb{R}^d\times\mathbb{R}^d\rightarrow \mathbb{R}^d$ combine the endpoint nodes representations with the shortest paths representation to get a final link representation. Since we consider undirected graphs, we consider $\mathcal{P}^*_G\{u, v\}:=\mathcal{P}^*_G(u, v)\cup \mathcal{P}^*_G(v, u)$. \fra{For disconnected pairs, $d_G(u,v)=+\infty$ and no shortest path exists. In this case, we use the endpoint-only sequence $(u,v)$ as a fallback input to the path encoder, with no additional no-path token. This sequence is not a graph path: the path branch uses only the two endpoint GNN embeddings, without intermediate
nodes.}

Although shortest paths provide a focused and efficient structural summary, they may lose information carried by alternative routes or larger subgraphs, reducing robustness and emphasizing local rather than global structure. 
We quantify this trade-off experimentally in Section \ref{subsec:ablation}, showing that replacing full path embeddings with simple distance information leads to significant performance degradation.

For a graph with \(n\) nodes and \(m\) edges, single-source BFS \citep{cormen2009introduction} runs in \(O(n+m)\) time on an adjacency-list representation. Although computing all-pairs shortest paths (APSP) would require \(O(n(n+m))\) time, SP4LP does not perform APSP.
In our experiments, we compute shortest paths only for the supervised train/validation/test links, which requires at most a small number of BFS traversals. Let $S$ be the number of distinct source nodes among these links; the preprocessing cost is therefore \(O(S\dot(n + m))\), typically several orders of magnitude smaller than \(O(n(n+m))\).
The resulting paths are cached and reused throughout training and inference, so no graph traversal is performed during model training or batched evaluation.
In fully-inductive settings where new nodes may appear, one can compute shortest-path trees by running a single BFS per new node, providing distances and parent pointers to all existing nodes at once. This preserves the same per-link inference efficiency and avoids BFS operations per queried link. A more detailed analysis of the overall complexity, including a comparison with prior approaches, is provided in Appendix~\ref{app:complexity}. In practice, all shortest paths are precomputed before training, and SP4LP simply retrieves the stored sequences during training and inference. No BFS operations are performed online.


\our is a general and flexible framework: both the underlying GNN used to compute node representations and the sequence model used to process the embeddings along the shortest path can be chosen modularly. For instance, the GNN component can be instantiated with architectures such as GCN \citep{kipf2016semi}, GAT \citep{velickovic2018graph} or GraphSAGE \citep{hamilton2017inductive}, while the sequence model can range from simple aggregation functions like injective summation as the one proposed in \citet{xu2019powerful}, to more complex architectures such as LSTMs \citep{hochreiter1997long}, GRU \citep{chung2014empirical} or Transformers \citep{vaswani2017attention}. An overview of \our is illustrated in Figure~\ref{fig:overview}.

The additional structural context given by the sequence of embeddings of nodes within the shortest path enables the model to distinguish links that are otherwise indistinguishable to standard message-passing methods, such as those involving automorphic nodes.

\begin{proposition}\label{prop}
\our does not suffer from the automorphic node problem.  
\end{proposition}
\fra{
The proof constructs two non-automorphic links with automorphic corresponding endpoints. Pure endpoint-based GNN models assign them identical representations, whereas their different shortest-path lengths allow a suitable instantiation of \our to distinguish them.
The complete proof is provided in Appendix~\ref{app:proofs}.
} As an example of non-automorphic links composed of automorphic nodes that \our can successfully distinguish, consider the links \((v, u)\) and \((v, u')\) shown in Figure~\ref{fig:myfigure}. \fra{The two links have automorphic corresponding endpoints: $v\simeq v$ via the identity and $u\simeq u'$ via a feature-preserving graph automorphism. However, the links $(v,u)$ and $(v,u')$ are not automorphic, since their endpoints have different shortest-path distances.}
This asymmetry is captured by the distinct shortest paths between the endpoints: the shortest path from $v$ to $u'$ consists of $v$, and orange node, and $u'$, whereas the shortest path from $v$ to $u$ includes $v$, and orange node, a yellow node, another orange node, and finally $u$.
\fra{
This example also defines an interpretable synthetic classification task: distinguish the same-side pair $(v,u')$ from the cross-graph pair $(v,u)$ across randomly relabeled copies of the graph, preserving node features. A pure endpoint-based GNN assigns identical representations to the two pairs and therefore cannot distinguish their classes. In contrast, a suitable instantiation of \our can separate them through their different shortest-path lengths. This is a representational comparison, rather than an empirical evaluation of training or generalization.
}
In addition to overcoming this limitation, \our is strictly more expressive than several state-of-the-art message passing methods for link representation learning.
\fra{The comparison with NCN assumes that the queried edge is removed, if present, and that both models use the resulting graph for message passing and structural feature extraction. It concerns the general \our formulation aggregating all shortest paths, rather than the single-path implementation used in our experiments.}

\begin{theorem}\label{thm}
\fra{
\our is strictly more expressive than pure GNN link representation models, NCN, the set-based horizontal GDGNN link encoder, and the degree-free vertical GDGNN link encoder. Moreover, neither Neo-GNN nor BUDDY is more expressive than \our.
}
\end{theorem}

\fra{
For each baseline covered by the strict expressiveness claim, the proof first shows how an appropriate choice of $\phi$,
$\mathrm{AGG}$, and $\rho$ reproduces its link representation under the stated assumptions. Strictness follows from examples of links that the baseline cannot distinguish but \our separates using path length, internal-node information, or node order.
The non-domination result for Neo-GNN and BUDDY follows from the NCN comparison by transitivity. The complete proof is provided in Appendix~\ref{app:proofs}. } In the following sections, we complement the theoretical analysis with an extensive experimental evaluation, showing that \our also achieves state-of-the-art performance on standard link prediction benchmarks.

\subsection{Further comparison with NCN}
\label{sec:comparisonncn}
NCN represents a link by aggregating the GNN embeddings of its endpoints and their common neighbors, making it particularly vulnerable to graph incompleteness. When two endpoints share no neighbors, NCN reduces to a pure GNN and can no longer leverage structural information, a far from rare situation in practice: the fraction of positive test links without common neighbors reaches 46\%, 55\%, and 67\% on Cora, Citeseer, and Pubmed, and 10\%, 37\%, and 52\% on ogbl-ppa, ogbl-citation2, and ogbl-collab (only ogbl-ddi is unaffected, at 0.04\%).
In contrast, \our\ encodes the sequence of node embeddings along the shortest path between endpoints, which is always defined in connected graphs and captures richer structural patterns even in sparse or incomplete settings. Moreover, while NCN pools neighbors as unordered sets, \our\ encodes paths as ordered sequences via LSTMs or Transformers, yielding strictly higher expressiveness (Theorem~\ref{thm}).

\section{Experiments}
\label{sec:exp}

We evaluate \our on real-world link prediction benchmarks, where observed edges are treated as positive examples and non-edges, sampled via negative sampling, as negatives. We consider three Planetoid citation networks, Cora, Citeseer, and Pubmed~\citep{yang2016revisiting}, and four OGB datasets, ogbl-citation2, ogbl-ppa, ogbl-collab and ogbl-ddi~\citep{hu2020open}. For the Planetoid datasets, we use a single fixed split across all experiments. Dataset statistics are reported in Table~\ref{app:tab:dataset-stats} of \cref{app:sec:dataset-statas}.

As baseline methods we consider \textbf{Pure GNN methods}: Graph Convolutional Network (GCN)~\citep{kipf2016semi}, Graph Attention Network (GAT)~\citep{veličković2018graph}, GraphSAGE~\citep{hamilton2017inductive}, and Graph Autoencoder (GAE)~\citep{kipf2016variational}, and \textbf{Structural Features GNN methods}: SEAL~\citep{zhang2018link}, BUDDY~\citep{chamberlaingraph}, Neo-GNN~\citep{yun2021neo}, NBFNet~\citep{zhu2021neural}, NCN~\citep{wangneural}, NCNC~\citep{wangneural} and  PEG~\citep{wangequivariant}, GDGNN~\citep{kong2022geodesic},
MPLP~\citep{dong2024pure}.
\revision{We additionally evaluate two recent link prediction models that fall outside the message-passing taxonomy of Section~\ref{sec:related-work}: LPFormer~\citep{shomer2024lpformer}, a transformer-based model with attention over node-pair contexts, and GraphGPT~\citep{zhao2025graphgpt}, a generative graph transformer that linearizes graphs via Eulerian paths.}


As described in Section~\ref{sec:method}, \our is a general framework that supports different choices of GNN backbone and sequence model $\phi$ (Equation~\ref{eq:our}), both treated as hyperparameters and tuned on the validation set.
We explore GCN, GAT, and GraphSAGE as backbones, and LSTM, Transformer, and an injective sum aggregator~\citep{xu2019powerful} as sequence models. We use an MLP for $\rho$ and, for efficiency, select the first shortest path returned by BFS as AGG.
Appendix~\ref{app:exp} reports implementation details, shortest-path computation, and hyperparameter configurations. Code is available at\footnote{\url{https://github.com/francescoferrini/sp4lp}}

\begin{table*}[!t]
\footnotesize
\setlength{\tabcolsep}{2pt}
\centering
\scriptsize
\caption{MRR and Hits@K (\%) on citation (Cora, Citeseer, Pubmed) and OGB benchmarks, following the HeaRT evaluation setting~\cite{li2023evaluating}. Hits@10 is reported for citation datasets and Hits@20 for OGB. Top three results per metric are highlighted as \first{first}, \second{second}, \third{third}. \textit{OOM} = out of memory; \textit{$>$24h} = did not complete within 24 hours. Methods are grouped into pure GNN baselines (\textbf{GNN}), GNNs augmented with structural features (\textbf{SF GNN}), and graph transformers (\textbf{GT}). Standard deviations over 5 runs are in \cref{app:sec:addiotional-res}. Note: GDGNN's original paper reports OGB results under a different protocol that does not follow HeaRT (which uses harder negative sampling and is significantly more expensive).}
\label{tab:res-main-merged}
\begin{tabular}{p{4pt}lcccccccccccccc}
\toprule
&\textbf{Models}
& \multicolumn{2}{c}{\textbf{Cora}}
& \multicolumn{2}{c}{\textbf{Citeseer}}
& \multicolumn{2}{c}{\textbf{Pubmed}}
& \multicolumn{2}{c}{\textbf{Ogbl-ddi}}
& \multicolumn{2}{c}{\textbf{Ogbl-collab}}
& \multicolumn{2}{c}{\textbf{Ogbl-ppa}}
& \multicolumn{2}{c}{\textbf{Ogbl-citation2}} \\
\cmidrule(lr){3-4}\cmidrule(lr){5-6}\cmidrule(lr){7-8}\cmidrule(lr){9-10}\cmidrule(lr){11-12}\cmidrule(lr){13-14}\cmidrule(lr){15-16}
& & MRR & H@10 & MRR & H@10 & MRR & H@10 & MRR & H@20 & MRR & H@20 & MRR & H@20 & MRR & H@20 \\
\midrule
\multirow{4}{*}{\rotatebox{90}{GNN}}
&GCN  & 16.61 & 36.26 & 21.09 & 47.23 & 7.13 & 15.22 & \third{13.46} & 64.76 & 6.09 & \third{22.48} & 26.94 & 68.38 & 19.98 & 51.72 \\
&GAT  & 13.84 & 32.89 & 19.58 & 45.30 & 4.95 & 9.99 & 12.92 & \second{66.83} & 4.18 & 18.30 & \textit{OOM} & \textit{OOM} & \textit{OOM} & \textit{OOM} \\
&SAGE & 14.74 & 34.65 & 21.09 & 48.75 & \third{9.40} & \third{20.54} & 12.60 & \first{67.19} & 5.53 & 21.26 & 27.27 & 69.49 & 22.05 & 53.13 \\
&GAE  & \second{18.32} & \second{37.95} & 25.25 & 49.65 & 5.27 & 10.50 & 3.49 & 17.81 & \textit{OOM} & \textit{OOM} & \textit{OOM} & \textit{OOM} & \textit{OOM} & \textit{OOM} \\
\midrule
\multirow{2}{*}{\rotatebox{90}{GT}}
&LPFORMER & \third{16.80} & 34.03 & \third{26.34} & \third{51.72} & \second{9.99} & \second{21.43} & 13.20 & 62.66 & \second{7.62} & 21.04 & \second{40.25} & \second{84.10} & \second{24.70} & \first{57.30} \\
&GraphGPT & 8.07 & 18.03 & 8.72 & 19.78 & 3.18 & 5.98 & \textit{$>$24h} & \textit{$>$24h} & 1.73 & 6.24 & \textit{$>$24h} & \textit{$>$24h} & \textit{$>$24h} & \textit{$>$24h} \\
\midrule
\multirow{9}{*}{\rotatebox{90}{SF GNN}}
&GDGNN    & 6.32 & 15.97 & 7.93 & 14.25 & 0.96 & 1.10 & \textit{$>$24h} & \textit{$>$24h} & \textit{$>$24h} & \textit{$>$24h} & \textit{$>$24h} & \textit{$>$24h} & \textit{$>$24h} & \textit{$>$24h} \\
&SEAL     & 10.67 & 24.27 & 13.16 & 27.37 & 5.88 & 12.47 & 9.99 & 49.74 & 6.43 & 21.57 & 29.71 & 76.77 & 20.60 & 48.62 \\
&BUDDY    & 13.71 & 30.40 & 22.84 & 48.35 & 7.56 & 16.78 & 12.43 & 58.71 & 5.67 & \second{23.35} & 27.70 & 71.50 & 19.17 & 47.81 \\
&Neo-GNN  & 13.95 & 31.27 & 17.34 & 41.74 & 7.74 & 17.88 & 10.86 & 51.94 & 5.23 & 21.03 & 21.68 & 64.81 & 16.12 & 43.17 \\
&NCN      & 14.66 & 35.14 & \second{28.65} & \second{53.41} & 5.84 & 13.22 & 12.86 & \third{65.82} & 5.09 & 20.84 & 35.06 & 81.89 & \third{23.35} & 53.76 \\
&NCNC     & 14.98 & \third{36.70} & 24.10 & \first{53.72} & 8.58 & 18.81 & \textit{$>$24h} & \textit{$>$24h} & 4.73 & 20.49 & 33.52 & \third{82.24} & 19.61 & 51.69 \\
&NBFNet   & 13.56 & 31.12 & 14.29 & 31.39 & \textit{$>$24h} & \textit{$>$24h} & \textit{$>$24h} & \textit{$>$24h} & \textit{OOM} & \textit{OOM} & \textit{OOM} & \textit{OOM} & \textit{OOM} & \textit{OOM} \\
&PEG      & 15.73 & 36.03 & 21.01 & 45.56 & 4.40 & 8.70 & 12.05 & 50.12 & 4.83 & 18.29 & \textit{OOM} & \textit{OOM} & \textit{OOM} & \textit{OOM} \\
&MPLP     & 12.57 & 27.70 & 12.55 & 27.33 & 5.19 & 10.97 & \second{13.94} & 59.17 & \third{6.79} & \first{25.10} & \first{41.40} & \first{84.80} & 23.11 & \second{55.51} \\

\cmidrule(l{2pt}r{2pt}){2-16}
&\our\xspace{\scriptsize (our)}
& \first{28.84} & \first{45.39}
& \first{29.48} & 49.52
& \first{10.87} & \first{23.01}
& \first{15.00} & 47.96
& \first{9.46} & 20.00
& \third{36.45} & 76.90
& \first{24.91} & \third{55.45} \\
\bottomrule
\end{tabular}
\end{table*}

\paragraph{Evaluation Setting} We evaluate model performance under the more challenging and realistic HeaRT evaluation setting~\cite{li2023evaluating}. In this setting, each positive target link (i.e., an existing link) is ranked against a carefully selected set of hard negative samples (i.e., non-existing links), providing a more realistic assessment of link prediction performance in practical scenarios. We adopt two standard ranking metrics: Hits@K and Mean Reciprocal Rank (MRR). Following the HeaRT protocol, we report Hits@10 and MRR for Cora, Citeseer, and Pubmed, and Hits@20 along with MRR for ogbl-collab, ogbl-ddi, ogbl-ppa and ogbl-citation2. The same set of negative samples is used across all positive links, as specified in the HeaRT benchmark. The HeaRT evaluation introduces significantly harder negative samples compared to traditional evaluation settings, resulting in a more challenging and realistic benchmark. \citet{li2023evaluating} show that this leads to a substantial performance drop across most models, with GNNs specifically designed for link prediction often being outperformed by simple heuristics or general-purpose GNNs. By adopting this challenging setting, we ensure a rigorous and meaningful comparison of model performance under conditions that closely resemble real-world applications.

Table~\ref{tab:res-main-merged} presents the performance of \our and the baseline 
models in terms of MRR and Hits@10 on Cora, Citeseer, and Pubmed, and MRR and 
Hits@20 on the OGB datasets. \our ranks first in terms of MRR on six out of seven 
datasets and third on the remaining one. The improvements in MRR are often 
substantial: on Cora, for instance, \our achieves a 57\% gain over the second-best 
method, GAE, and on Ogbl-Collab a 24\% gain over the second-best method, LPFormer, 
highlighting a markedly stronger empirical predictive capability of our approach. 
\our also achieves the best Hits@K score on Cora and Pubmed, and ranks among the 
top three on Ogbl-Citation2. On Ogbl-Collab, although \our does not rank among the 
top performers on Hits@20, it remains comparable to several baselines (e.g., NCN, 
NCNC, Neo-GNN) when accounting for standard deviations 
(Appendix~\ref{app:sec:addiotional-res}). \fra{MRR and Hits@K capture different aspects of ranking quality:
moving a positive link from rank 10 to rank 2 improves MRR without changing Hits@20. Thus, the MRR gains of \our need not translate into higher top-20 coverage. On ogbl-ddi, shortest paths have average length approximately two, and only 0.04\% of positive test links lack common neighbors, limiting the additional structural information provided by a single path. On ogbl-collab and ogbl-ppa, selecting one shortest path may likewise discard evidence distributed across multiple paths or common neighbors. Greater theoretical expressiveness therefore does not guarantee improvements on every metric: empirical performance also depends on graph structure, optimization, and architectural inductive bias.} In addition to 
achieving the best performance in several datasets, \our is also the most 
consistent model across all benchmarks. Unlike previous state-of-the-art models 
such as BUDDY and Neo-GNN, which tend to struggle on datasets like Cora, Citeseer 
and Pubmed, \our maintains strong performance regardless of dataset 
characteristics. Overall, these results clearly demonstrate the superiority of 
\our under the more challenging and realistic HeaRT evaluation setting, confirming 
its effectiveness for real-world link prediction tasks.

\subsection{Ablation Study}\label{subsec:ablation}


\begin{wraptable}{l}{0.56\columnwidth}
\vspace{-8pt}
\centering
\caption{Ablation study results (\%). MRR and Hits@K are reported for two model variants: (1) \textit{Sequence Model Only}, and (2) \textit{GNN + Shortest Path Length}. The full model \our consistently outperforms both variants.}
\label{tab:ablation}
\scriptsize
\begin{tabular}{lcccccc}
\toprule
\textbf{Models} 
& \multicolumn{2}{c}{\textbf{Cora}} 
& \multicolumn{2}{c}{\textbf{Citeseer}} 
& \multicolumn{2}{c}{\textbf{Pubmed}} \\
 & MRR & H@10 & MRR & H@10 & MRR & H@10\\
\midrule
GNN + SP 
& 8.32 & 20.02 
& 21.13 & 45.05 
& 8.52 & 15.32\\
Seq. Model 
& 16.86 & 36.03 
& 10.93 & 19.23 
& 7.12 & 13.56 \\
\midrule
\our 
& \textbf{28.84} & \textbf{45.39} 
& \textbf{29.48} & \textbf{49.52} 
& \textbf{10.87} & \textbf{23.01} \\
\bottomrule
\end{tabular}
\vspace{-10pt}
\end{wraptable}
We perform an ablation study to evaluate the contribution of the main components of our model. In particular, we investigate two simplified variants to understand the importance of node representation learning and sequential modeling along the shortest path between target nodes. \textbf{(1) Sequence Model Only}: in this variant, the sequential model operates directly on the raw input features of the nodes along the shortest path, without incorporating node representations learned by the GNN. This setup isolates the contribution of the sequential model in capturing relational patterns based solely on node features and structural path information. \textbf{(2) GNN + Shortest Path Length:} in this variant, the sequential model is completely removed. Link prediction is performed using only the learned node representations from the GNN, combined with the length of the shortest path between the target nodes. This evaluates the effectiveness of combining node embeddings with simple distance information, without explicitly modeling the intermediate nodes along the path.

Table~\ref{tab:ablation} reports the results of these ablations on Cora,  Citeseer, and Pubmed. Both variants show a clear performance drop compared to the  full model, demonstrating the importance of jointly leveraging node representations  and sequential modeling. Note that, since both variants inherit the GNN configuration  optimized for the full \our pipeline rather than being independently retuned, their  absolute scores are not directly comparable to the standalone GNN baselines in  Table~\ref{tab:res-main-merged}; the meaningful comparison is between each variant and the full \our. The relative ranking of the two variants is dataset-dependent:  Sequence Model Only performs better on Cora, while GNN + Shortest Path Length is  stronger on Citeseer and Pubmed. In all cases, neither variant approaches the full  \our: replacing the path embedding with its length discards the information carried  by node representations along the path, while operating on raw features deprives  the model of the broader structural context implicitly encoded by message passing.  The full \our consistently achieves the best results across all datasets, confirming  the importance of combining learned node representations with sequential modeling  over the shortest paths.

\fra{
The ablation results also suggest that joint optimization of the GNN and sequence encoder contributes to the variability observed on the smaller datasets. On Cora, the MRR standard deviation increases from 0.28 for GNN + shortest-path length and 1.26 for the sequence-only
model to 5.45 for the full model. On Citeseer, the corresponding values are 0.99, 2.93, and 10.17. Hits@10 exhibits the same pattern (Appendix~\ref{app:sec:addiotional-res}, Table~5). These observations suggest an interaction between the two components, rather than establishing a definitive cause of the variability.
}
\subsection{Scalability Analysis}

\begin{figure}[t]
    \centering
    \includegraphics[width=\linewidth]{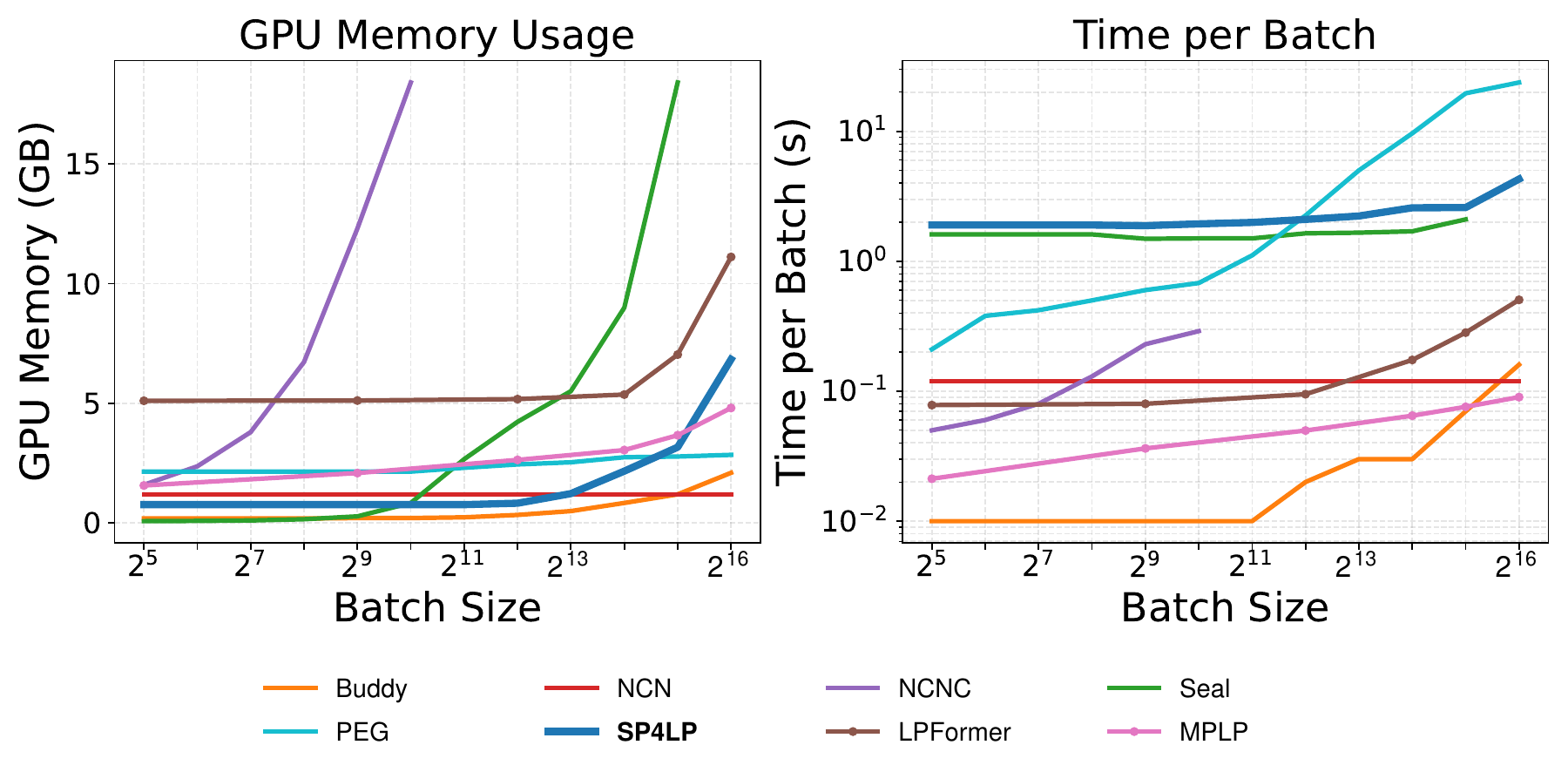}
    \caption{Inference time and GPU memory usage on ogbl-collab, measured during the prediction of a single batch of test links. \fra{GDGNN is omitted because it did not complete within 24 hours. GraphGPT is omitted because its autoregressive decoding over variable-length sequences is not directly comparable to the batch-forward inference measured here.}}
    \label{fig:scalability}
\end{figure}

We assess the scalability of \our by examining how its GPU memory consumption and inference time evolve as the batch size increases, in comparison to several baseline methods. The results are presented in Figure~\ref{fig:scalability}. 
In terms of GPU memory usage, \our exhibits remarkable efficiency: memory consumption remains nearly constant across small to medium batch sizes, and increases moderately only for the largest batches, starting with 0.77 GB and reaching at most 6.84 GB while consistently avoiding out-of-memory (OOM) failures. 

PEG also maintains low memory usage; however, this advantage is undermined by its impractically slow inference, limiting its applicability in large-scale scenarios. SEAL, while competitive with \our in terms of inference speed, suffers from excessive memory consumption, rapidly exceeding 18 GB and encountering OOM issues beyond a batch size of 32,768. NCNC is even more constrained, experiencing OOM failures already at relatively small batch sizes. Considering inference time, \our matches the efficiency of SEAL, requiring only 2.1 seconds for o batch size of 16,384 and scaling smoothly to 4.29 seconds at 65,536. PEG, by contrast, is significantly slower, already taking 5 seconds at a batch size of 8,192 and exceeding 23 seconds at the maximum batch size tested. Although NCN and Buddy consistently achieve low inference times, this comes at the cost of substantially lower predictive performance, as they rely on simple heuristics with limited modeling capacity. This trade-off is clearly reflected in the results of Table~\ref{tab:res-main-merged}. \fra{LPFormer and MPLP achieve lower inference times than \our at the measured batch sizes. At batch size 65,536, their GPU memory usage is 11.12 GB and 4.81 GB, respectively, compared with 6.84 GB for \our. These results show a trade-off: \our achieves higher MRR
on ogbl-collab, while these baselines offer faster inference.}



\section{Conclusion}\label{conc}
We introduced \our, a message-passing framework for link representation that enhances the expressiveness of standard GNNs by incorporating sequential modeling over the shortest path between target nodes. Following the GNN-then-SF paradigm, \our combines the efficiency of computing node embeddings only once with high expressive power, explicitly modeling multi-hop relational patterns via a sequence encoder applied to node embeddings along the shortest path. We formally proved that \our is strictly more expressive than several state-of-the-art link representation models, and extensive experiments under the HeaRT evaluation protocol confirm that it achieves state-of-the-art performance across diverse datasets.

\textbf{Limitations.} \our presents two main limitations. First, it requires a preprocessing step to compute and cache shortest paths, which adds complexity compared to fully end-to-end models, although this cost is amortized across training and inference. Second, \our's advantage is reduced on graphs with very short average shortest-path length (e.g., ogbl-ddi, with average length $\approx 2$), where path-based modeling has limited room to exploit structural information beyond what local methods already capture.

\textbf{Future work.} We plan to extend \our in several directions: incorporating multiple (or all) shortest paths to improve robustness and capture richer structural signals; adapting \our to heterogeneous graphs~\citep{ferrini2025a}, where shortest paths remain well-defined and the path encoder can naturally incorporate node and edge type embeddings; \fra{and extending our preliminary analysis of truncated BFS
to larger graphs and other approximation techniques, such as landmark-based methods, to reduce preprocessing cost.}

\clearpage
\section*{Acknowledgments}
FF, BL, and AP were funded by the European Union under Grant Agreement No. 101120763 (TANGO), and Grant Agreement No. 101120237 (ELIAS).  FF and  BL were funded by Horizon Europe Grant Agreement No. 101297599 (EURO-3C). AL and VL were funded by the Research Council of Norway through its Centre of Excellence Integreat \textendash{} The Norwegian Centre for Knowledge-Driven Machine Learning (project no. 332645). Views and opinions expressed are however those of the author(s) only and do not necessarily reflect those of the European Union or the European Health and Digital Executive Agency (HaDEA). Neither the European Union nor the granting authorities can be held responsible for them.

\bibliographystyle{neurips2026}
\bibliography{bibliography}

@inproceedings{
airale2025simple,
title={Simple Path Structural Encoding for Graph Transformers},
author={Louis Airale and Antonio Longa and Mattia Rigon and Andrea Passerini and Roberto Passerone},
booktitle={Forty-second International Conference on Machine Learning},
year={2025},
url={https://openreview.net/forum?id=t3zwUqibMq}
}

@article{lu2011link,
  title={Link prediction in complex networks: A survey},
  author={L{\"u}, Linyuan and Zhou, Tao},
  journal={Physica A: statistical mechanics and its applications},
  volume={390},
  number={6},
  pages={1150--1170},
  year={2011},
  publisher={Elsevier}
}

@inproceedings{cheng2025edge,
  title={Edge Classification on Graphs: New Directions in Topological Imbalance},
  author={Cheng, Xueqi and Wang, Yu and Liu, Yunchao and Zhao, Yuying and Aggarwal, Charu C and Derr, Tyler},
  booktitle={Proceedings of the Eighteenth ACM International Conference on Web Search and Data Mining},
  pages={392--400},
  year={2025}
}

@inproceedings{lichter2025computational,
  title={Computational Complexity of the Weisfeiler-Leman Dimension},
  author={Lichter, Moritz and Ra{\ss}mann, Simon and Schweitzer, Pascal},
  booktitle={33rd EACSL Annual Conference on Computer Science Logic (CSL 2025)},
  pages={13--1},
  year={2025},
  organization={Schloss Dagstuhl--Leibniz-Zentrum f{\"u}r Informatik}
}

@article{cai1992optimal,
  title={An optimal lower bound on the number of variables for graph identification},
  author={Cai, Jin-Yi and F{\"u}rer, Martin and Immerman, Neil},
  journal={Combinatorica},
  volume={12},
  number={4},
  pages={389--410},
  year={1992},
  publisher={Springer}
}

@article{dawar2020generalizations,
  title={Generalizations of k-dimensional Weisfeiler--Leman stabilization},
  author={Dawar, Anuj and Vagnozzi, Danny},
  journal={Moscow Journal of Combinatorics and Number Theory},
  volume={9},
  number={3},
  pages={229--252},
  year={2020},
  publisher={Mathematical Sciences Publishers}
}

@inproceedings{dong2019link,
  title={The link regression problem in graph streams},
  author={Dong, Bowen and Aggarwal, Charu C and Philip, S Yu},
  booktitle={2019 IEEE International Conference on Big Data (Big Data)},
  pages={1088--1095},
  year={2019},
  organization={IEEE}
}

@article{zhou2021progresses,
  title={Progresses and challenges in link prediction},
  author={Zhou, Tao},
  journal={Iscience},
  volume={24},
  number={11},
  year={2021},
  publisher={Elsevier}
}

@inproceedings{NEURIPS2025_b1cd7227,
 author = {Lachi, Veronica and Ferrini, Francesco and Longa, Antonio and Lepri, Bruno and Passerini, Andrea and Jaeger, Manfred},
 booktitle = {Advances in Neural Information Processing Systems},
 doi = {10.52202/085713-4090},
 editor = {D. Belgrave and C. Zhang and H. Lin and R. Pascanu and P. Koniusz and M. Ghassemi and N. Chen},
 pages = {122617--122646},
 publisher = {Curran Associates, Inc.},
 title = {Bridging Theory and Practice in Link Representation with Graph Neural Networks},
 url = {https://proceedings.neurips.cc/paper_files/paper/2025/file/b1cd72276feff8173462e3f733ac66f8-Paper-Conference.pdf},
 volume = {38, Main Conference},
 year = {2025}
}

@article{mckay2014practical,
  title   = {Practical graph isomorphism, {II}},
  author  = {McKay, Brendan D. and Piperno, Adolfo},
  journal = {Journal of Symbolic Computation},
  volume  = {60},
  pages   = {94--112},
  year    = {2014}
}

@article{
ferrini2025a,
title={A Self-Explainable Heterogeneous {GNN} for Relational Deep Learning},
author={Francesco Ferrini and Antonio Longa and Andrea Passerini and Manfred Jaeger},
journal={Transactions on Machine Learning Research},
issn={2835-8856},
year={2025},
url={https://openreview.net/forum?id=8Q4qxe9a9Z},
note={}
}

@inproceedings{
lachi2024a,
title={A Simple and Expressive Graph Neural Network Based Method for Structural Link Representation},
author={Veronica Lachi and Francesco Ferrini and Antonio Longa and Bruno Lepri and Andrea Passerini},
booktitle={ICML 2024 Workshop on Geometry-grounded Representation Learning and Generative Modeling},
year={2024},
url={https://openreview.net/forum?id=EGGSCLyVrz}
}

@article{srinivasan2019equivalence,
  title={On the equivalence between positional node embeddings and structural graph representations},
  author={Srinivasan, Balasubramaniam and Ribeiro, Bruno},
  journal={arXiv preprint arXiv:1910.00452},
  year={2019}
}

@article{zhang2021labeling,
  title={Labeling trick: A theory of using graph neural networks for multi-node representation learning},
  author={Zhang, Muhan and Li, Pan and Xia, Yinglong and Wang, Kai and Jin, Long},
  journal={Advances in Neural Information Processing Systems},
  volume={34},
  pages={9061--9073},
  year={2021}
}

@article{kipf2016semi,
  title={Semi-supervised classification with graph convolutional networks},
  author={Kipf, Thomas N and Welling, Max},
  journal={arXiv preprint arXiv:1609.02907},
  year={2016}
}

@article{kipf2016variational,
  title={Variational graph auto-encoders},
  author={Kipf, Thomas N and Welling, Max},
  journal={arXiv preprint arXiv:1611.07308},
  year={2016}
}

@misc{xu2019powerful,
      title={How Powerful are Graph Neural Networks?}, 
      author={Keyulu Xu and Weihua Hu and Jure Leskovec and Stefanie Jegelka},
      year={2019},
      eprint={1810.00826},
      archivePrefix={arXiv},
      primaryClass={cs.LG}
}

@article{zhu2021neural,
  title={Neural bellman-ford networks: A general graph neural network framework for link prediction},
  author={Zhu, Zhaocheng and Zhang, Zuobai and Xhonneux, Louis-Pascal and Tang, Jian},
  journal={Advances in neural information processing systems},
  volume={34},
  pages={29476--29490},
  year={2021}
}

@inproceedings{li2023evaluating,
title={Evaluating Graph Neural Networks for Link Prediction: Current Pitfalls and New Benchmarking},
author={Juanhui Li and Harry Shomer and Haitao Mao and Shenglai Zeng and Yao Ma and Neil Shah and Jiliang Tang and Dawei Yin},
booktitle={Thirty-seventh Conference on Neural Information Processing Systems Datasets and Benchmarks Track},
year={2023},
url={https://openreview.net/forum?id=YdjWXrdOTh}
}

@article{yun2021neo,
  title={Neo-gnns: Neighborhood overlap-aware graph neural networks for link prediction},
  author={Yun, Seongjun and Kim, Seoyoon and Lee, Junhyun and Kang, Jaewoo and Kim, Hyunwoo J},
  journal={Advances in Neural Information Processing Systems},
  volume={34},
  pages={13683--13694},
  year={2021}
}

@inproceedings{chamberlaingraph,
title={Graph Neural Networks for Link Prediction with Subgraph Sketching},
author={Benjamin Paul Chamberlain and Sergey Shirobokov and Emanuele Rossi and Fabrizio Frasca and Thomas Markovich and Nils Yannick Hammerla and Michael M. Bronstein and Max Hansmire},
booktitle={The Eleventh International Conference on Learning Representations },
year={2023},
url={https://openreview.net/forum?id=m1oqEOAozQU}
}

@inproceedings{wangequivariant,
title={Equivariant and Stable Positional Encoding for More Powerful Graph Neural Networks},
author={Haorui Wang and Haoteng Yin and Muhan Zhang and Pan Li},
booktitle={International Conference on Learning Representations},
year={2022},
url={https://openreview.net/forum?id=e95i1IHcWj}
}

@inproceedings{wangneural,
title={Neural Common Neighbor with Completion for Link Prediction},
author={Xiyuan Wang and Haotong Yang and Muhan Zhang},
booktitle={The Twelfth International Conference on Learning Representations},
year={2024},
url={https://openreview.net/forum?id=sNFLN3itAd}
}

@inproceedings{
velickovic2018graph,
title={Graph Attention Networks},
author={Petar Veli{\v{c}}kovi{\'{c}} and Guillem Cucurull and Arantxa Casanova and Adriana Romero and Pietro Liò and Yoshua Bengio},
booktitle={International Conference on Learning Representations},
year={2018},
url={https://openreview.net/forum?id=rJXMpikCZ},
}

@article{zhang2018link,
  title={Link prediction based on graph neural networks},
  author={Zhang, Muhan and Chen, Yixin},
  journal={Advances in neural information processing systems},
  volume={31},
  year={2018}
}

@inproceedings{morris2019weisfeiler,
  title={Weisfeiler and leman go neural: Higher-order graph neural networks},
  author={Morris, Christopher and Ritzert, Martin and Fey, Matthias and Hamilton, William L and Lenssen, Jan Eric and Rattan, Gaurav and Grohe, Martin},
  booktitle={Proceedings of the AAAI conference on artificial intelligence},
  volume={33},
  pages={4602--4609},
  year={2019}
}

@article{jha2022prediction,
  title={Prediction of protein--protein interaction using graph neural networks},
  author={Jha, Kanchan and Saha, Sriparna and Singh, Hiteshi},
  journal={Scientific Reports},
  volume={12},
  number={1},
  pages={8360},
  year={2022},
  publisher={Nature Publishing Group UK London}
}

@article{nickel2015review,
  title={A review of relational machine learning for knowledge graphs},
  author={Nickel, Maximilian and Murphy, Kevin and Tresp, Volker and Gabrilovich, Evgeniy},
  journal={Proceedings of the IEEE},
  volume={104},
  number={1},
  pages={11--33},
  year={2015},
  publisher={IEEE}
}

@inproceedings{ying2018graph,
  title={Graph convolutional neural networks for web-scale recommender systems},
  author={Ying, Rex and He, Ruining and Chen, Kaifeng and Eksombatchai, Pong and Hamilton, William L and Leskovec, Jure},
  booktitle={Proceedings of the 24th ACM SIGKDD international conference on knowledge discovery \& data mining},
  pages={974--983},
  year={2018}
}

@article{rossi2021knowledge,
  title={Knowledge graph embedding for link prediction: A comparative analysis},
  author={Rossi, Andrea and Barbosa, Denilson and Firmani, Donatella and Matinata, Antonio and Merialdo, Paolo},
  journal={ACM Transactions on Knowledge Discovery from Data (TKDD)},
  volume={15},
  number={2},
  pages={1--49},
  year={2021},
  publisher={ACM New York, NY, USA}
}

@inproceedings{wang2021apan,
  title={Apan: Asynchronous propagation attention network for real-time temporal graph embedding},
  author={Wang, Xuhong and Lyu, Ding and Li, Mengjian and Xia, Yang and Yang, Qi and Wang, Xinwen and Wang, Xinguang and Cui, Ping and Yang, Yupu and Sun, Bowen and others},
  booktitle={Proceedings of the 2021 international conference on management of data},
  pages={2628--2638},
  year={2021}
}

@article{liang2025line,
  title={Line graph neural networks for link weight prediction},
  author={Liang, Jinbi and Pu, Cunlai and Shu, Xiangbo and Xia, Yongxiang and Xia, Chengyi},
  journal={Physica A: Statistical Mechanics and its Applications},
  pages={130406},
  year={2025},
  publisher={Elsevier}
}

@inproceedings{shomer2024lpformer,
  title={LPFormer: an adaptive graph transformer for link prediction},
  author={Shomer, Harry and Ma, Yao and Mao, Haitao and Li, Juanhui and Wu, Bo and Tang, Jiliang},
  booktitle={Proceedings of the 30th ACM SIGKDD Conference on Knowledge Discovery and Data Mining},
  pages={2686--2698},
  year={2024}
}

@inproceedings{borgwardt2005shortest,
  title={Shortest-path kernels on graphs},
  author={Borgwardt, Karsten M and Kriegel, Hans-Peter},
  booktitle={Fifth IEEE international conference on data mining (ICDM'05)},
  pages={8--pp},
  year={2005},
  organization={IEEE}
}

@article{ying2021transformers,
  title={Do transformers really perform badly for graph representation?},
  author={Ying, Chengxuan and Cai, Tianle and Luo, Shengjie and Zheng, Shuxin and Ke, Guolin and He, Di and Shen, Yanming and Liu, Tie-Yan},
  journal={Advances in neural information processing systems},
  volume={34},
  pages={28877--28888},
  year={2021}
}

@book{cormen2009introduction,
  title        = {Introduction to Algorithms},
  author       = {Cormen, Thomas H. and Leiserson, Charles E. and Rivest, Ronald L. and Stein, Clifford},
  year         = {2009},
  edition      = {3rd},
  publisher    = {MIT Press},
  address      = {Cambridge, MA}
}

@article{hamilton2017inductive,
  title={Inductive representation learning on large graphs},
  author={Hamilton, Will and Ying, Zhitao and Leskovec, Jure},
  journal={Advances in neural information processing systems},
  volume={30},
  year={2017}
}

@article{chung2014empirical,
  title={Empirical evaluation of gated recurrent neural networks on sequence modeling},
  author={Chung, Junyoung and Gulcehre, Caglar and Cho, KyungHyun and Bengio, Yoshua},
  journal={arXiv preprint arXiv:1412.3555},
  year={2014}
}

@article{vaswani2017attention,
  title={Attention is all you need},
  author={Vaswani, Ashish and Shazeer, Noam and Parmar, Niki and Uszkoreit, Jakob and Jones, Llion and Gomez, Aidan N and Kaiser, {\L}ukasz and Polosukhin, Illia},
  journal={Advances in neural information processing systems},
  volume={30},
  year={2017}
}

@article{hochreiter1997long,
  title={Long short-term memory},
  author={Hochreiter, Sepp and Schmidhuber, J{\"u}rgen},
  journal={Neural computation},
  volume={9},
  number={8},
  pages={1735--1780},
  year={1997},
  publisher={MIT press}
}

@inproceedings{veličković2018graph,
title={Graph Attention Networks},
author={Petar Veli{\v{c}}kovi{\'{c}} and Guillem Cucurull and Arantxa Casanova and Adriana Romero and Pietro Liò and Yoshua Bengio},
booktitle={International Conference on Learning Representations},
year={2018},
url={https://openreview.net/forum?id=rJXMpikCZ},
}

@inproceedings{yang2016revisiting,
  title={Revisiting semi-supervised learning with graph embeddings},
  author={Yang, Zhilin and Cohen, William and Salakhudinov, Ruslan},
  booktitle={International conference on machine learning},
  pages={40--48},
  year={2016},
  organization={PMLR}
}

@article{hu2020open,
  title={Open graph benchmark: Datasets for machine learning on graphs},
  author={Hu, Weihua and Fey, Matthias and Zitnik, Marinka and Dong, Yuxiao and Ren, Hongyu and Liu, Bowen and Catasta, Michele and Leskovec, Jure},
  journal={Advances in neural information processing systems},
  volume={33},
  pages={22118--22133},
  year={2020}
}

@techreport{hagberg2008exploring,
  title={Exploring network structure, dynamics, and function using NetworkX},
  author={Hagberg, Aric and Swart, Pieter J and Schult, Daniel A},
  year={2008},
  institution={Los Alamos National Laboratory (LANL), Los Alamos, NM (United States)}
}

@article{sommer2014shortest,
  title={Shortest-path queries in static networks},
  author={Sommer, Christian},
  journal={ACM Computing Surveys (CSUR)},
  volume={46},
  number={4},
  pages={1--31},
  year={2014},
  publisher={ACM New York, NY, USA}
}

@article{
tola2025proxi,
title={{PROXI}: Challenging the {GNN}s for Link Prediction},
author={Astrit Tola and Jack Myrick and Baris Coskunuzer},
journal={Transactions on Machine Learning Research},
issn={2835-8856},
year={2025},
url={https://openreview.net/forum?id=u9EHndbiVw},
note={}
}

@article{kong2022geodesic,
  title={Geodesic graph neural network for efficient graph representation learning},
  author={Kong, Lecheng and Chen, Yixin and Zhang, Muhan},
  journal={Advances in neural information processing systems},
  volume={35},
  pages={5896--5909},
  year={2022}
}

@inproceedings{
dong2024pure,
title={Pure Message Passing Can Estimate Common Neighbor for Link Prediction},
author={Kaiwen Dong and Zhichun Guo and Nitesh V Chawla},
booktitle={The Thirty-eighth Annual Conference on Neural Information Processing Systems},
year={2024},
url={https://openreview.net/forum?id=Xa3dVaolKo}
}

@inproceedings{
zhao2025graphgpt,
title={Graph{GPT}: Generative Pre-trained Graph Eulerian Transformer},
author={Qifang Zhao and Weidong Ren and Tianyu Li and Hong Liu and Xingsheng He and Xiaoxiao Xu},
booktitle={Forty-second International Conference on Machine Learning},
year={2025},
url={https://openreview.net/forum?id=4RdzeucFmW}
}

@inproceedings{chen2020convolutional,
  title={Convolutional kernel networks for graph-structured data},
  author={Chen, Dexiong and Jacob, Laurent and Mairal, Julien},
  booktitle={international conference on machine learning},
  pages={1576--1586},
  year={2020},
  organization={PMLR}
}

@inproceedings{michel2023path,
  title={Path neural networks: Expressive and accurate graph neural networks},
  author={Michel, Gaspard and Nikolentzos, Giannis and Lutzeyer, Johannes F and Vazirgiannis, Michalis},
  booktitle={International Conference on Machine Learning},
  pages={24737--24755},
  year={2023},
  organization={PMLR}
}

@article{chen2024learning,
  title={Learning long range dependencies on graphs via random walks},
  author={Chen, Dexiong and Schulz, Till Hendrik and Borgwardt, Karsten},
  journal={arXiv preprint arXiv:2406.03386},
  year={2024}
}

\newpage
\appendix

\section{Proofs}\label{app:proofs} 

\textbf{Proposition \ref{prop}} \our does not suffer from the automorphic node problem.
\begin{proof}
\fra{
According to Proposition~\ref{prop:auto}, a model class $M$
suffers from the \emph{automorphic node problem} if, for every
graph $G=(V,E,\mathbf{X}^0)$ and every pair of non-automorphic
links $(u,v)$ and $(u',v')$ with $u\simeq u'$ and $v\simeq v'$,
all $F\in M$ satisfy
\[
F((u,v),G)=F((u',v'),G).
\]
To show that \our does not suffer from this problem, it suffices
to exhibit one such graph and pair of links, and one
instantiation $F\in\our$, for which
\[
F((u,v),G)\neq F((u',v'),G).
\]
}
Such an example is provided in Figure~\ref{fig:myfigure}: the shortest path from $v$ to $u'$ consists of $v$, an orange node, and $u'$, whereas the shortest path from $v$ to $u$ includes $v$, an orange node, a yellow node, another orange node, and finally $u$. Thus, simply using a summation as function $\phi$, leads to distinct representations for links $(v,u)$ and $(v,u')$.
\end{proof}

\begin{figure}
    \centering
    \includegraphics[width=0.3\linewidth]{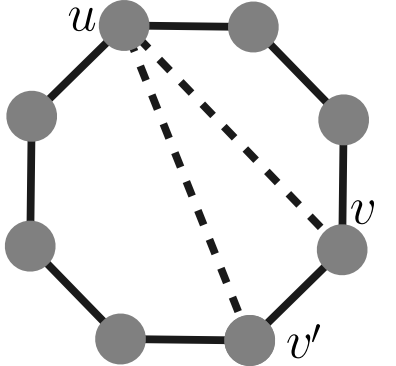}
    \caption{\fra{An eight-node cycle with constant node features. Dashed lines indicate queried pairs, not graph edges. Both $(u,v)$ and $(u,v')$ have no common neighbors, but their shortest-path distances are three and four, respectively. NCN assigns them identical representations, whereas a length-sensitive instantiation of \our distinguishes them.}}
    \label{fig:reg}
\end{figure}

\textbf{Theorem \ref{thm}}
\fra{
\our is strictly more expressive than pure GNN link representation models, NCN, the set-based horizontal GDGNN link encoder, and the degree-free vertical GDGNN link encoder. Moreover, neither Neo-GNN nor BUDDY is more expressive than \our.
}
\begin{proof} We proceed by prove each comparison separately.
\begin{itemize}
    \item \underline{\our is strictly more expressive than Pure GNN.}

We prove this by noting that \our architecture (Equation~\ref{eq:our}) generalizes that of pure GNNs, meaning that \our can simulate any pure GNN by simply ignoring the shortest path information. Thus, for any pair of non-automorphic links that a specific pure GNN can distinguish, there exists a configuration of \our that distinguishes them as well. We can conclude that \our is strictly more expressive than pure GNNs considering as example of pair of links indistinguishable by GNNs but distinguishable by \our the one provided in the proof of Proposition~\ref{prop}.

\item \underline{\our is strictly more expressive than NCN.}

\fra{
We first establish that \our can reproduce NCN in the target-link-removed setting. Let $H$ denote the graph used by both models for message passing and structural feature extraction, with the queried edge removed if present. Write $\mathbf{h}_w=\gnn(w,H)$.

If $u$ and $v$ have common neighbors in $H$, their shortest paths have length two and are precisely the paths $(u,w,v)$ with $w\in N_H(u)\cap N_H(v)$. Choose $\phi$ to return $\mathbf{h}_w/2$ for each such three-node sequence and zero for sequences of any other length, including the disconnected-pair fallback. With summation as $\text{AGG}$, this yields
\[
\sum_{w\in N_H(u)\cap N_H(v)}\mathbf{h}_w.
\]
The factor $1/2$ accounts for the two orientations of each path included in $\mathcal{P}^*_H\{u,v\}$. When there are no common neighbors, the aggregate is zero, as in NCN. Choosing $\rho$ to reproduce NCN's combination of endpoint embeddings and the common-neighbor aggregate therefore recovers its link representation. This establishes $\mathrm{NCN}\preceq\our$ under the stated setting.
}

\fra{
To establish strictness, consider the eight-node cycle with constant node features shown in Figure~\ref{fig:reg}, with $d_H(u,v)=3$ and $d_H(u,v')=4$. The cycle is vertex-transitive, so all nodes receive the same embedding from a message-passing GNN. Neither queried pair
has common neighbors; consequently, NCN assigns the same representation to $(u,v)$ and $(u,v')$. The two links are not automorphic, since automorphisms preserve shortest-path distances. In contrast, their shortest-path sequences contain four and five nodes, respectively.
Choose a nonzero constant node embedding $\mathbf{x}$, let $\phi$ sum the embeddings within each path, and let $\text{AGG}$ average the resulting path representations. The two path representations are then $4\mathbf{x}$ and $5\mathbf{x}$, which a suitable $\rho$ distinguishes. Together with the inclusion established above, this proves strictness in the target-link-removed setting.
}



\item \fra{\underline{Neo-GNN and BUDDY do not dominate \our.}}

\fra{
Theorem 2 of \citet{wangneural} establishes that neither Neo-GNN nor BUDDY is more expressive than NCN. Since $\mathrm{NCN}\preceq\our$, if either model $M$ could subsume \our, transitivity would imply
\[
\mathrm{NCN}\preceq\our\preceq M,
\]
contradicting that result. Therefore, neither Neo-GNN nor BUDDY is more expressive than \our. This does not establish that \our is strictly more expressive than either model.
}

\item \underline{\our is strictly more expressive than GDGNN.}

\paragraph{ (Horizontal GDGNN).}
We show that \our is strictly more expressive than the set-based horizontal GDGNN link encoder.
First, SP4LP subsumes horizontal GDGNN: choose the SP4LP path encoder $\phi$ to be permutation-invariant over the nodes
on a shortest path (i.e., $\phi(\mathbf{h}_P)=\mathrm{POOL}(\{\mathbf{h}_x:x\in P\})$), and choose $\mathrm{AGG}$ to return the
representation of the same deterministically selected shortest path $\pi(u,v)$ used by horizontal GDGNN (e.g., the path returned by a fixed
shortest-path routine with a fixed tie-breaking rule). With the same decoder, the resulting SP4LP equals the horizontal GDGNN on every link query.
To prove strictness, consider two non-automorphic links $(u,v)$ and $(u',v')$ each having a \emph{unique} shortest path,
$P=(u,a,b,v)$ and $P'=(u',b,a,v')$, with embeddings satisfying $\mathbf{h}_u=\mathbf{h}_{u'}$, $\mathbf{h}_v=\mathbf{h}_{v'}$, and
$\mathbf{h}_a\neq \mathbf{h}_b$ (e.g., due to distinct node features). Any set-based horizontal GDGNN aggregates over the multiset
$\{\mathbf{h}_u,\mathbf{h}_a,\mathbf{h}_b,\mathbf{h}_v\}$ and hence produces identical pooled features for $P$ and $P'$; with matching endpoints,
it outputs identical link representations and cannot distinguish the two queries. In contrast, SP4LP can choose an order-sensitive sequence encoder
$\phi$ (e.g., an RNN/Transformer) such that $\phi(\mathbf{h}_u,\mathbf{h}_a,\mathbf{h}_b,\mathbf{h}_v)\neq
\phi(\mathbf{h}_{u'},\mathbf{h}_b,\mathbf{h}_a,\mathbf{h}_{v'})$, and since each query has a unique shortest path, $\mathrm{AGG}$ is trivial.
Thus SP4LP distinguishes a pair of non-automorphic link queries that horizontal GDGNN cannot, proving strictness.

\paragraph{ (Vertical GDGNN).}
We show that SP4LP is strictly more expressive than the vertical GDGNN link encoder.
For subsumption, define the SP4LP path encoder $\phi$ to extract only the first and last internal nodes of a shortest path:
for $P=(u,x_1,\dots,x_{\ell-1},v)$ let $\phi(\mathbf{h}_P)=[\mathbf{h}_{x_1};\mathbf{h}_{x_{\ell-1}}]$, and let $\mathrm{AGG}$ pool these
two positions across all shortest paths (with the same permutation-invariant $\mathrm{POOL}$ as vertical GDGNN), yielding exactly the pooled feature
over the vertical endpoint-neighbor set $S^{\mathrm{V}}(u,v)=(N(u)\cap W(u,v))\cup (N(v)\cap W(u,v))$, where $W(u,v)=\{w:\ d(u,w)+d(w,v)=d(u,v)\}$.
Using the same decoder $\rho$, SP4LP matches the degree-free vertical GDGNN on every link query.
For strictness, consider two non-automorphic link queries $(u,v)$ and $(u',v')$ with unique shortest paths
$P=(u,a,x,b,v)$ and $P'=(u',a,y,b,v')$, where the first/last internal nodes coincide ($a$ and $b$) but the middle internal node differs ($x\neq y$),
and assume $\mathbf{h}_u=\mathbf{h}_{u'}$, $\mathbf{h}_v=\mathbf{h}_{v'}$, and $\mathbf{h}_x\neq \mathbf{h}_y$ (e.g., via distinct features).
Since each query has a unique shortest path, the degree-free vertical GDGNN uses the same vertical set $S^{\mathrm{V}}(u,v)=\{a,b\}=S^{\mathrm{V}}(u',v')$,
hence produces identical pooled features and cannot distinguish the two queries. In contrast, SP4LP may use an order-sensitive $\phi$ on the full path sequence,
so that $\phi(\mathbf{h}_u,\mathbf{h}_a,\mathbf{h}_x,\mathbf{h}_b,\mathbf{h}_v)\neq \phi(\mathbf{h}_{u'},\mathbf{h}_a,\mathbf{h}_y,\mathbf{h}_b,\mathbf{h}_{v'})$
whenever $\mathbf{h}_x\neq \mathbf{h}_y$, and $\mathrm{AGG}$ is trivial under uniqueness of the shortest path. Therefore, SP4LP distinguishes a pair of non-automorphic
link queries that the degree-free vertical GDGNN cannot, proving strictness.

\paragraph{Remark (Scope).}
The strictness result applies to the \emph{degree-free} vertical GDGNN class. If the vertical model is augmented with additional induced-subgraph structural features (e.g., degree-in-induced-subgraph), the inclusion and strictness relationships may change and require a separate analysis.

\end{itemize}

\end{proof}

\section{Additional results}\label{app:sec:addiotional-res}

We complement the main results of Section \ref{sec:exp} with additional tables reporting the standard deviation computed over five different random seeds, to better assess the stability of each method.

\paragraph{Real-world Datasets: Results with Standard Deviations}
Table~\ref{app:tab:mrr-std} and Table~\ref{app:tab:hits-std} expand the main results in Table~\ref{tab:res-main-merged} by including both the mean and standard deviation of MRR and Hits@$K$ across runs.

\begin{table}[ht]
\centering
\resizebox{\textwidth}{!}{%
\begin{tabular}{p{4pt}lccccccc}
\toprule
&Models & Cora & Citeseer & Pubmed & Ogbl-ddi & Ogbl-collab & Ogbl-ppa & Ogbl-Citation2\\
& &  MRR & MRR & MRR & MRR & MRR & MRR & MRR \\
\midrule
\multirow{4}{*}{\rotatebox{90}{GNN}}
&GCN & \meanstd{16.61}{0.30} & \meanstd{21.09}{0.88} & \meanstd{7.13}{0.27} & \third{\meanstd{13.46}{0.34}} & \meanstd{6.09}{0.38} &\meanstd{26.94}{0.48} & \meanstd{19.98}{0.35}\\
&GAT & \meanstd{13.84}{0.68} & \meanstd{19.58}{0.84} & \meanstd{4.95}{0.14} & \meanstd{12.92}{0.39} & \meanstd{4.18}{0.33} & OOM & OOM\\
&SAGE & \meanstd{14.74}{0.69} & \meanstd{21.09}{1.15} & \third{\meanstd{9.40}{0.70}} & \meanstd{12.60}{0.72} & \meanstd{5.53}{0.50} &\meanstd{27.27}{0.30} & \meanstd{22.05}{0.12}\\
&GAE & \second{\meanstd{18.32}{0.41}} & \meanstd{25.25}{0.82} & \meanstd{5.27}{0.25} & \meanstd{3.49}{1.73} & OOM & OOM & OOM\\
\midrule

\multirow{2}{*}{\rotatebox{90}{GT}}
&LPFORMER & \third{\meanstd{16.80}{0.52}} & \third{\meanstd{26.34}{0.67}} & \second{\meanstd{9.99}{0.52}}  & \meanstd{13.20}{0.54}  & \second{\meanstd{7.62}{0.26}}  & \second{\meanstd{40.25}{0.24}} & \second{\meanstd{24.70}{0.55}} \\
&GraphGPT & \meanstd{8.07}{2.98} & \meanstd{8.72}{3.21} & \meanstd{3.18}{3.45} & $>$24h & \meanstd{1.73}{3.91} & $>$24h & $>$24h \\
\midrule
\multirow{9}{*}{\rotatebox{90}{SF GNN}}
&GDGNN & \meanstd{6.32}{0.17} & \meanstd{7.93}{0.45} & \meanstd{0.96}{0.02} & $>$24h & $>$24h & $>$24h & $>$24h\\
&SEAL & \meanstd{10.67}{3.46} & \meanstd{13.16}{1.66} & \meanstd{5.88}{0.53} & \meanstd{9.99}{0.90} & \meanstd{6.43}{0.32} & \meanstd{29.71}{0.71} & \meanstd{20.60}{1.28}\\
&BUDDY & \meanstd{13.71}{0.59} & \meanstd{22.84}{0.36} & \meanstd{7.56}{0.18} & \meanstd{12.43}{0.50} & \meanstd{5.67}{0.36} & \meanstd{27.70}{0.33} & \meanstd{19.17}{0.10}\\
&Neo-GNN & \meanstd{13.95}{0.39} & \meanstd{17.34}{0.84} & \meanstd{7.74}{0.30} & \meanstd{10.86}{2.16} & \meanstd{5.23}{0.90} & \meanstd{21.68}{1.14} & \meanstd{16.12}{0.25}\\
&NCN & \meanstd{14.66}{0.95} & \second{\meanstd{28.65}{1.21}} & \meanstd{5.84}{0.22} & \meanstd{12.86}{0.78} & \meanstd{5.09}{0.38} & \meanstd{35.06}{0.26} & \third{\meanstd{23.35}{0.28}}\\
&NCNC & \meanstd{14.98}{1.00} & \meanstd{24.10}{0.65} & \meanstd{8.58}{0.59} & $>$24h & \meanstd{4.73}{0.86} & \meanstd{33.52}{0.26} & \meanstd{19.61}{0.54}\\
&NBFNet & \meanstd{13.56}{0.58} & \meanstd{14.29}{0.80} & $>$24h & $>$24h & OOM & OOM & OOM \\
&PEG & \meanstd{15.73}{0.39} & \meanstd{21.01}{0.77} & \meanstd{4.40}{0.41} & \meanstd{12.05}{1.14} & \meanstd{4.83}{0.21} & OOM & OOM \\
&MPLP & \meanstd{12.57}{0.23} & \meanstd{12.55}{0.17} & \meanstd{5.19}{0.09} & \second{\meanstd{13.94}{0.96}} & \third{\meanstd{6.79}{1.02}} & \first{\meanstd{41.40}{0.65}} & \meanstd{23.11}{1.76} \\

\cmidrule(l{2pt}r{2pt}){2-9}
&\our & \first{\meanstd{28.84}{5.45}} & \first{\meanstd{29.48}{10.17}} & \first{\meanstd{10.87}{0.31}} & \first{\meanstd{15.00}{0.57}} & \first{\meanstd{9.46}{0.55}} & \third{\meanstd{36.45}{1.21}} & \first{\meanstd{24.91}{0.41}}\\
\bottomrule
\end{tabular}
}
\caption{MRR results across all datasets, following the HeaRT evaluation setting~\cite{li2023evaluating}. The top three results for each metric are highlighted using \first{first}, \second{second}, and \third{third}. \textit{OOM} indicates that the model ran out of memory, while \textit{$>$24h} denotes that the method did not complete within 24 hours. Methods are grouped into pure GNN baselines (\textbf{GNN}), GNNs augmented with structural features (\textbf{SF GNN}), and graph transformers (\textbf{GT}).}\label{app:tab:mrr-std}
\end{table}

\begin{table}[ht]
\centering
\resizebox{\textwidth}{!}{%
\begin{tabular}{p{4pt}lccccccc}
\toprule
&Models & Cora & Citeseer & Pubmed & Ogbl-ddi & Ogbl-collab & Ogbl-ppa & Ogbl-Citation2\\
& & Hits@10 & Hits@10 & Hits@10 & Hits@20 & Hits@20 & Hits@20 & Hits@20\\
\midrule
\multirow{4}{*}{\rotatebox{90}{GNN}}
&GCN & \meanstd{36.26}{1.14} & \meanstd{47.23}{1.88} & \meanstd{15.22}{0.57} & \meanstd{64.76}{1.45} & \third{\meanstd{22.48}{0.81}} & \meanstd{68.38}{0.73} & \meanstd{51.72}{0.46} \\
&GAT & \meanstd{32.89}{1.27} & \meanstd{45.30}{1.30} & \meanstd{9.99}{0.64} & \second{\meanstd{66.83}{2.23}} & \meanstd{18.30}{1.42} & OOM & OOM \\
&SAGE & \meanstd{34.65}{1.47} & \meanstd{48.75}{1.85} & \third{\meanstd{20.54}{1.40}} & \first{\meanstd{67.19}{1.18}} & \meanstd{21.26}{1.32} & \meanstd{69.49}{0.43} & \meanstd{53.13}{0.15} \\
&GAE & \second{\meanstd{37.95}{1.24}} & \meanstd{49.65}{1.48} & \meanstd{10.50}{0.46} & \meanstd{17.81}{9.80} & OOM & OOM & OOM \\
\midrule
\multirow{2}{*}{\rotatebox{90}{GT}}
&LPFORMER & \meanstd{34.03}{1.33} & \third{\meanstd{51.72}{1.83}} & \second{\meanstd{21.43}{1.30}}  & \meanstd{62.66}{0.37}  & \meanstd{21.04}{0.16}  & \second{\meanstd{84.10}{0.10}} & \first{\meanstd{57.30}{0.50}} \\ 
&GraphGPT & \meanstd{18.03}{2.12} & \meanstd{19.78}{1.11} & \meanstd{5.98}{3.21} & $>$24h & \meanstd{6.24}{2.67} & $>$24h & $>$24h \\
\midrule
\multirow{9}{*}{\rotatebox{90}{SF GNN}}
&GDGNN & \meanstd{15.97}{0.31} & \meanstd{14.25}{1.15} & \meanstd{1.10}{0.02} & $>$24h & $>$24h & $>$24h & $>$24h \\
&SEAL & \meanstd{24.27}{6.74} & \meanstd{27.37}{3.20} & \meanstd{12.47}{1.23} & \meanstd{49.74}{2.39} & \meanstd{21.57}{0.38} & \meanstd{76.77}{0.94} & \meanstd{48.62}{1.93} \\
&BUDDY & \meanstd{30.40}{1.18} & \meanstd{48.35}{1.18} & \meanstd{16.78}{0.53} & \meanstd{58.71}{1.63} & \second{\meanstd{23.35}{0.73}} & \meanstd{71.50}{0.68} & \meanstd{47.81}{0.37} \\
&Neo-GNN & \meanstd{31.27}{0.72} & \meanstd{41.74}{1.18} & \meanstd{17.88}{0.71} & \meanstd{51.94}{10.33} & \meanstd{21.03}{3.39} & \meanstd{64.81}{2.26} & \meanstd{43.17}{0.53} \\
&NCN & \meanstd{35.14}{1.04} & \second{\meanstd{53.41}{1.46}} & \meanstd{13.22}{0.56} & \third{\meanstd{65.82}{2.66}} & \meanstd{20.84}{1.31} & \meanstd{81.89}{0.31} & \meanstd{53.76}{0.20} \\
&NCNC & \third{\meanstd{36.70}{1.57}} & \first{\meanstd{53.72}{0.97}} & \meanstd{18.81}{1.16} & $>$24h & \meanstd{20.49}{3.97} & \third{\meanstd{82.24}{0.40}} & \meanstd{51.69}{1.48} \\
&NBFNet & \meanstd{31.12}{0.75} & \meanstd{31.39}{1.34} & $>$24h & $>$24h & OOM & OOM & OOM \\
&PEG & \meanstd{36.03}{0.75} & \meanstd{45.56}{1.38} & \meanstd{8.70}{1.26} & \meanstd{50.12}{6.55} & \meanstd{18.29}{1.06} & OOM & OOM \\
&MPLP & \meanstd{27.70}{1.51} & \meanstd{27.33}{3.00} & \meanstd{10.97}{0.31} & \meanstd{59.17}{1.65} & \first{\meanstd{25.10}{1.01}} & \first{\meanstd{84.80}{0.98}} & \second{\meanstd{55.51}{1.12}} \\

\cmidrule(l{2pt}r{2pt}){2-9}
&\our & \first{\meanstd{45.39}{9.58}} & \meanstd{49.52}{9.77} & \first{\meanstd{23.01}{0.39}} & \meanstd{47.96}{3.82} & \meanstd{20.00}{1.20} & \meanstd{76.9}{1.11} & \third{\meanstd{55.45}{0.92}} \\
\bottomrule
\end{tabular}
}
\caption{Hits@K (\%) results across all datasets, following the HeaRT evaluation setting~\cite{li2023evaluating}. The top three results for each metric are highlighted using \first{first}, \second{second}, and \third{third}. \textit{OOM} indicates that the model ran out of memory, while \textit{$>$24h} denotes that the method did not complete within 24 hours. Methods are grouped into pure GNN baselines (\textbf{GNN}), GNNs augmented with structural features (\textbf{SF GNN}), and graph transformers (\textbf{GT}).}\label{app:tab:hits-std}
\end{table}

\paragraph{Ablation Study: Results with Standard Deviations}
Similarly, Table~\ref{app:ablation-std} complements the ablation results in Table~\ref{tab:ablation} by reporting mean and standard deviation for Cora, Citeseer, and Pubmed.

\begin{table}[ht]
\centering
\resizebox{\textwidth}{!}{%
\begin{tabular}{lcccccc}
\toprule
\textbf{Models} & \multicolumn{2}{c}{\textbf{Cora}} & \multicolumn{2}{c}{\textbf{Citeseer}} & \multicolumn{2}{c}{\textbf{Pubmed}} \\
 & MRR & Hits@10 & MRR & Hits@10 & MRR & Hits@10\\
\midrule
\textit{GNN + SP len.} & \meanstd{8.32}{0.28} & \meanstd{20.02}{0.67} & \meanstd{21.13}{0.99} & \meanstd{45.05}{1.24} & \meanstd{8.52}{0.01} & \meanstd{15.32}{0.66}\\
\textit{Sequence Model} & \meanstd{16.86}{1.26} & \meanstd{36.03}{1.75}	& \meanstd{10.93}{2.93} & \meanstd{19.23}{4.20}	& \meanstd{7.12}{0.75} &\meanstd{13.56}{0.85} \\
\midrule
\our & \textbf{\meanstd{28.84}{5.45}} & \textbf{\meanstd{45.39}{9.58}} & \textbf{\meanstd{29.48}{10.17}} & \textbf{\meanstd{49.52}{9.77}} & \textbf{\meanstd{10.87}{0.31}} & \textbf{\meanstd{23.01}{0.39}} \\
\bottomrule
\end{tabular}%
}
\caption{Ablation study results (\%). MRR and Hits@K with mean and std. deviations over 5 runs with different seeds.}
\label{app:ablation-std}
\end{table}

\paragraph{\fra{Truncated BFS.}}
\fra{
We evaluate the trade-off between predictive performance and preprocessing cost on Pubmed by limiting BFS to a maximum
number of hops. Pairs reachable within the cutoff retain their exact shortest path; otherwise, we use the endpoint-only fallback $(u,v)$. Table~\ref{tab:truncated-bfs} reports results from a single run. Increasing the cutoff improves predictive performance at the cost of longer preprocessing.
}

\begin{table}[ht]
    \centering
    \fra{
    \begin{tabular}{lrrr}
        \toprule
        Maximum hops & MRR (\%) & Hits@10 (\%) & Preprocessing (s) \\
        \midrule
        2    & 7.00  & 14.98 & 1   \\
        5    & 7.59  & 16.34 & 150 \\
        7    & 9.16  & 18.21 & 320 \\
        Full & 10.87 & 23.01 & 570 \\
        \bottomrule
    \end{tabular}
    }
    \caption{\fra{Predictive performance and shortest-path
    preprocessing time for truncated BFS on Pubmed.
    Full denotes BFS without a depth cutoff.
    Results are from a single run.}}
    \label{tab:truncated-bfs}
\end{table}

\paragraph{\fra{Graph symmetries and empirical gains.}}
\fra{ We quantify structural graph symmetries using $r'_G=1-(|\mathcal{O}_G|-1)/(|V_G|-1)$, where $\mathcal{O}_G$ is the set of node-automorphism orbits. We compute exact orbits using nauty~\citep{mckay2014practical} on the three citation datasets, for which this computation was feasible. Table~\ref{tab:automorphism-analysis} compares this measure with the MRR improvement over GCN. We observe no monotonic relationship between graph symmetry and the improvement of \our. Thus, automorphisms motivate pair-specific representations,
but do not alone explain empirical gains, which also depend on node features, path distributions, optimization, and architectural inductive biases.
}

\begin{table}[ht]
    \centering
    \fra{
    \begin{tabular}{lrrrr}
        \toprule
        Dataset & $r'_G$ & GCN MRR & \our MRR & $\Delta$ MRR \\
        \midrule
        Cora     & 0.127 & 16.61 & 28.84 & +12.23 \\
        Citeseer & 0.372 & 21.09 & 29.48 & +8.39  \\
        Pubmed   & 0.341 & 7.13  & 10.87 & +3.74  \\
        \bottomrule
    \end{tabular}
    }
    \caption{\fra{Structural graph symmetries and MRR gains
    over GCN. MRR is reported in percent and gains in
    percentage points.}}
    \label{tab:automorphism-analysis}
\end{table}

\section{Datasets statistics}\label{app:sec:dataset-statas}
Table~\ref{app:tab:dataset-stats} summarizes the main datasets used in our link prediction experiments. Cora, Citeseer, and Pubmed are well-known citation networks frequently used as benchmarks for graph-based learning methods. These datasets are relatively small, both in the number of nodes and edges. In contrast, the datasets from the Open Graph Benchmark (OGB), namely ogbl-collab and ogb-ddi, are substantially larger and more complex, offering challenging scenarios for evaluating model scalability and performance on large-scale graphs.

For all datasets, we adopt the splits provided by the \cite{li2023evaluating} setting.
\begin{table}[ht]
\setlength{\tabcolsep}{4pt}
\centering
\begin{tabular}{lccccccc}
\toprule
\textbf{Dataset} & Cora & Citeseer & Pubmed & ogbl-collab & ogbl-ddi  & ogbl-ppa & ogbl-citation2\\
\midrule
\#Nodes & 2708 & 3327 & 18717 & 235868 & 4267 & 576289 & 2927963\\
\#Edges & 5278 & 4676 & 44327 & 1285465 & 1334889 & 30326273 & 30561187\\
\bottomrule
\end{tabular}
\caption{Dataset statistics.}
\label{app:tab:dataset-stats}
\end{table}

\paragraph{\fra{Shortest-path statistics.}}
\fra{
Table~\ref{tab:path-statistics} reports shortest-path statistics
for each dataset and split, separately for positive and negative
queries. No-path percentages are computed over all queries
in the corresponding group; means and variances are computed
only over pairs connected by a finite path.
Disconnected pairs use the endpoint-only fallback described
in Section~4.
Such pairs are particularly frequent in the smaller citation
graphs, while their prevalence is generally low in the
validation and test splits of the larger benchmarks.
}
\fra{
Statistics use undirected, unweighted graphs.
Validation/test queries use the training-only graph.
For the three citation datasets, training statistics use
a \texttt{train\_val} subset sampled with seed 999 and
matched to the validation size; each queried positive edge
is removed before evaluating it and its associated negatives.
For OGB training statistics, 70\% of shuffled training-edge
occurrences form the message-passing graph and the remaining
30\% provide positive queries. Repeated pairs can occur
in both subsets, yielding distance-one positive queries.
}

\begin{table}[p]
    \centering
    \footnotesize
    \setlength{\tabcolsep}{4pt}
    \fra{
    \begin{tabular}{lllrrrrr}
        \toprule
        Dataset & Split & Queries & Total & No path
        & No path (\%) & Mean & Variance \\
        \midrule
        Cora & Train$^\dagger$ & Pos. & 263 & 35
             & 13.31 & 3.00 & 2.13 \\
             & Train$^\dagger$ & Neg. & 131500 & 30077
             & 22.87 & 6.87 & 5.14 \\
             & Val. & Pos. & 263 & 41
             & 15.59 & 3.09 & 2.71 \\
             & Val. & Neg. & 131500 & 18828
             & 14.32 & 5.48 & 5.33 \\
             & Test & Pos. & 527 & 77
             & 14.61 & 3.15 & 2.74 \\
             & Test & Neg. & 263500 & 36857
             & 13.99 & 5.48 & 5.20 \\
        \midrule
        Citeseer & Train$^\dagger$ & Pos. & 227 & 77
                 & 33.92 & 3.18 & 6.80 \\
                 & Train$^\dagger$ & Neg. & 113500 & 78813
                 & 69.44 & 9.77 & 12.93 \\
                 & Val. & Pos. & 227 & 77
                 & 33.92 & 3.12 & 4.16 \\
                 & Val. & Neg. & 113500 & 66483
                 & 58.58 & 7.64 & 17.27 \\
                 & Test & Pos. & 455 & 163
                 & 35.82 & 3.13 & 6.31 \\
                 & Test & Neg. & 227500 & 129764
                 & 57.04 & 7.58 & 17.59 \\
        \midrule
        Pubmed & Train$^\dagger$ & Pos. & 2216 & 549
               & 24.77 & 3.10 & 1.89 \\
               & Train$^\dagger$ & Neg. & 1108000 & 175019
               & 15.80 & 6.62 & 2.47 \\
               & Val. & Pos. & 2216 & 512
               & 23.10 & 3.05 & 1.63 \\
               & Val. & Neg. & 1108000 & 171651
               & 15.49 & 4.34 & 5.38 \\
               & Test & Pos. & 4432 & 1091
               & 24.62 & 3.04 & 1.51 \\
               & Test & Neg. & 2216000 & 364645
               & 16.46 & 4.53 & 5.53 \\
        \midrule
        ogbl-ddi & Train & Pos. & 320374 & 30
                 & 0.00936 & 2.00 & 0.00228 \\
                 & Train & Neg. & 320374 & 3759
                 & 1.17 & 2.31 & 0.25 \\
                 & Val. & Pos. & 1000 & 0
                 & 0 & 2.00 & 0 \\
                 & Val. & Neg. & 500000 & 0
                 & 0 & 2.00010 & 0.000100 \\
                 & Test & Pos. & 1000 & 0
                 & 0 & 2.00 & 0 \\
                 & Test & Neg. & 500000 & 0
                 & 0 & 2.00019 & 0.000190 \\
        \midrule
        ogbl-collab & Train & Pos. & 353716 & 18239
                   & 5.16 & 1.81 & 0.42 \\
                   & Train & Neg. & 353716 & 48568
                   & 13.73 & 7.30 & 3.07 \\
                   & Val. & Pos. & 60084 & 1111
                   & 1.85 & 2.17 & 2.10 \\
                   & Val. & Neg. & 30042000 & 531855
                   & 1.77 & 3.08 & 4.53 \\
                   & Test & Pos. & 46329 & 944
                   & 2.04 & 2.50 & 2.39 \\
                   & Test & Neg. & 23164500 & 420641
                   & 1.82 & 3.05 & 4.06 \\
        \midrule
        ogbl-ppa & Train & Pos. & 6369580 & 4652
                & 0.0730 & 2.11 & 0.12 \\
                & Train & Neg. & 6369580 & 82591
                & 1.30 & 4.39 & 0.63 \\
                & Val. & Pos. & 100000 & 0
                & 0 & 2.06 & 0.06 \\
                & Val. & Neg. & 50000000 & 47
                & 0.000094 & 3.66 & 0.73 \\
                & Test & Pos. & 100000 & 4
                & 0.004 & 2.06 & 0.06 \\
                & Test & Neg. & 50000000 & 1057
                & 0.002114 & 3.66 & 0.73 \\
        \midrule
        ogbl-citation2 & Train & Pos. & 9116399 & 97250
                      & 1.07 & 2.39 & 0.59 \\
                      & Train & Neg. & 9116399 & 514395
                      & 5.64 & 6.41 & 1.53 \\
                      & Val. & Pos. & 86596 & 69
                      & 0.0797 & 2.29 & 0.41 \\
                      & Val. & Neg. & 43298000 & 19457
                      & 0.0449 & 4.78 & 3.30 \\
                      & Test & Pos. & 86596 & 77
                      & 0.0889 & 2.29 & 0.40 \\
                      & Test & Neg. & 43298000 & 21591
                      & 0.0499 & 4.74 & 3.47 \\
        \bottomrule
    \end{tabular}
    }
    \caption{\fra{
    Shortest-path statistics for the analyzed query sets.
    Total and No path report the number of queries and
    disconnected queries, respectively.
    Means and population variances are computed over finite
    shortest-path distances only; the endpoint-only fallback
    is excluded. Distances are measured in hops.
    $^\dagger$For Cora, Citeseer, and Pubmed, training statistics
    refer to the sampled \texttt{train\_val} subset and its
    associated negative queries, rather than the full training set.
    }}
    \label{tab:path-statistics}
\end{table}

\section{Experimental Settings}\label{app:exp}
This section outlines the experimental setup used to evaluate all models. We describe the computational resources and the hyperparameter search space. Moreover for \our we include details regarding how the calculation of the shortest path is performed. Details are reported below.
\paragraph{Computational Resources} All experiments were conducted on a workstation running Ubuntu 22.04 with an AMD Ryzen 9 7950X CPU (32 threads), 124GB of RAM, and two NVIDIA GeForce RTX 4090 GPUs (24GB each).
\paragraph{Hyperparameters} All models are tuned using a grid search over learning rate $\in [1 \times 10^{-2}, 1 \times 10^{-3}]$, dropout $\in [0, 0.7]$, weight decay $\in [0, 10^{-4}, 10^{-7}]$, number of GNN layers $\in \{1, 2, 3\}$, hidden dimensions $\in \{32, 64, 128, 256\}$ and prediction layers $\in \{1, 2, 3\}$. For large-scale datasets, we follow the reduced search space adopted in~\citet{li2023evaluating} to avoid excessive compute. For \our, we additionally explore the choice of GNN component $\in \{\text{GCN}, \text{GraphSAGE}, \text{GAT}\}$ and sequence model $\in \{\text{LSTM}, \text{Transformer}\}$, the best models are shown in Table~\ref{tab:best-gnn-seq-models}. The best hyperparameters are selected based on validation performance. All reported metrics are averaged over 5 different seeds.

\begin{table}[ht]
    \centering
    \resizebox{\linewidth}{!}{%
    \fra{%
    \begin{tabular}{lccccccc}
        \toprule
        Parameter & Cora & Citeseer & Pubmed
        & ogbl-collab & ogbl-ddi & ogbl-ppa & ogbl-citation2 \\
        \midrule
        GNN & GCN & GCN & SAGE & SAGE & GCN & GCN & GCN \\
        Learning rate
        & 0.001 & 0.01 & 0.001 & 0.001 & 0.01 & 0.001 & 0.01 \\
        Dropout
        & 0.5 & 0.5 & 0.1 & 0 & 0 & 0.3 & 0.3 \\
        Weight decay
        & 0 & $10^{-7}$ & 0 & 0 & 0 & 0 & 0 \\
        GNN layers
        & 1 & 1 & 2 & 3 & 3 & 3 & 3 \\
        GNN hidden dimension
        & 128 & 128 & 128 & 256 & 256 & 256 & 128 \\
        Predictor layers
        & 3 & 2 & 3 & 3 & 3 & 3 & 3 \\
        Batch size
        & 1024 & 1024 & 1024 & 65536 & 65536 & 65536 & 65536 \\
        \midrule
        Sequence encoder
        & Transformer & LSTM & Transformer & Transformer
        & Transformer & Transformer & Transformer \\
        Sequence layers
        & 2 & 2 & 2 & 2 & 2 & 2 & 2 \\
        Attention heads
        & 4 & -- & 4 & 4 & 4 & 4 & 4 \\
        Path hidden dimension
        & 128 & 128 & 128 & 64 & 64 & 64 & 64 \\
        Path dropout
        & 0.1 & 0.1 & 0.1 & 0.2 & 0.2 & 0.2 & 0.2 \\
        Predictor hidden dimension
        & 256 & 256 & 256 & 64 & 64 & 64 & 64 \\
        \bottomrule
    \end{tabular}%
    }%
    }
    \caption{\fra{Selected hyperparameters for the reported
    \our results. SAGE denotes GraphSAGE; attention heads
    are not applicable to the LSTM encoder.}}
    \label{tab:best-gnn-seq-models}
\end{table}
\paragraph{Shortest path calculation for \our}

To compute shortest paths for \our, we use the \texttt{networkx.shortest\_path} BFS implementation~\citep{hagberg2008exploring}. Importantly, we compute and cache shortest paths only for the train/validation/test links provided by the HeaRT benchmark splits. \fra{During training, message passing and shortest-path extraction use a subset of training edges disjoint from the supervised positive edges. For validation and testing, both use the training-only graph, without adding validation or test edges.} \fra{A single BFS can serve multiple queries sharing the same source node, provided that they use the same path-extraction
graph. Queries evaluated on different edge-masked graphs
require separate computations.} \fra{The preprocessing cost is amortized through caching. Table~\ref{tab:truncated-bfs} reports the trade-off between
preprocessing time and predictive performance on Pubmed.}
The cached paths are reused across all training epochs and inference batches. No all-pairs shortest path computation is performed.
\fra{
When multiple shortest paths exist, we select the first path returned by BFS under a fixed adjacency ordering and cache it, keeping the selection deterministic throughout training and evaluation. To assess sensitivity to this choice, we fixed a trained GCN--Transformer model on Cora and varied only the paths used at test time. For each test link, we sampled up to ten alternative shortest paths, using repetition when fewer alternatives were available. Across these path selections, the standard deviations were 0.10 percentage points for MRR and 0.01 percentage points for Hits@10, indicating low sensitivity in this experiment.
}

In our implementation of \our, the sequential encoder can be instantiated as a Transformer with learnable positional encodings. We use PyTorch\textquoteright{}s TransformerEncoder, composed of TransformerEncoderLayer blocks with 4-head self-attention, feedforward sublayers, ReLU activation, and dropout. We tune the number of layers $\in \{1, 2\}$ , attention heads $\in \{2,3,4\}$, and feedforward dimensionality $\in \{32, 64,128\}$. A trainable positional embedding matrix is added to node embeddings to preserve path order. Variable-length paths are handled via a source key padding mask, and the output is aggregated using masked mean pooling followed by layer normalization. The resulting path representation is then combined with the GNN-derived embeddings of the source and target nodes to compute the final link score.

For the LSTM-based encoder, we implement both unidirectional and bidirectional variants using PyTorch\textquoteright{}s nn.LSTM. Input sequences are packed with pack\_padded\_sequence to handle variable-length paths. We tune the number of layers $\in \{1, 2\}$ and hidden size $\in \{32, 64, 128\}$. The final hidden state (or the concatenation of forward and backward states in the bidirectional case) is used as the path representation and combined with the GNN-based node embeddings for link prediction. Node embeddings are obtained via a GNN encoder selected from GCN, GAT, or GraphSAGE, depending on the experimental setting.

\paragraph{\fra{Reproducibility details.}}
\fra{
The precomputed \texttt{gnn\_feature} and \texttt{entity\_embedding} files used by the small-dataset
pipeline come from the HeaRT data release and are loaded using the same procedure as its baselines.
Training negatives follow the HeaRT baseline sampling procedure (\texttt{torch.randint}), without explicitly excluding held-out positive edges. Consequently, sampled training negatives may overlap
with held-out positives. For every \our run reported in Table~\ref{tab:res-main-merged}, all training, validation, and test paths, including paths for sampled training negatives, were recomputed from scratch with \texttt{--force\_recompute\_paths}, bypassing existing caches.
}

\section{Computational Complexity}
\label{app:complexity}
In this section, we analyze the computational complexity of our proposed \our model for link prediction, following the formalism and notation introduced in~\cite{wangneural}. This framework allows a direct comparison with prior approaches such as GAE, Neo-GNN, BUDDY, PEG, SEAL, NCN, and NCNC.

Let $n$ be the number of nodes, $m$ the number of edges, $\Delta$ the maximum degree, $d$ the dimensionality of node features or embeddings, $t$ the number of target links to predict, and $k$ the length of the shortest path between node pairs (typically $k \ll n$). We express total time complexity as $\mathcal{O}(B + C \cdot t)$, where $B$ is the precomputation cost (independent of $t$), and $C$ is the per-link cost.

Our \our method follows the GNN-then-Structural-Feature (SF) paradigm~\cite{wangneural}, applying a Message Passing Neural Network once across the entire graph to compute node embeddings, then leveraging shortest path extraction combined with sequence modeling for each candidate link.

Following the paradigm, the precomputation cost of \our is:
\[
B = \mathcal{O}(n\Delta d + nd^2 + T_{\text{sp}})
\]
where: $n\Delta d$ accounts for neighborhood aggregation in the GNN, $nd^2$ represents linear transformations in GNN layers, $T_{\text{sp}}$ is the cost of computing shortest paths (e.g., via BFS).

For sparse graphs, single-source BFS runs in $O(m)$ time. Although all-pairs shortest paths (APSP) would cost $O(nm)$, \our never computes APSP. In all benchmark settings, we compute shortest paths only for the supervised train/validation/test links. Let $S$ be the number of distinct endpoint nodes among these links. The total preprocessing cost is therefore
\[
T_{\mathrm{sp}} = O(S \cdot m),
\]
typically several orders of magnitude smaller than $O(nm)$.
Once these paths are cached, no BFS is performed during training or batched inference. In fully inductive scenarios, shortest-path trees for new nodes can be obtained via a single BFS per new node, which provides distances and parent pointers to all existing nodes at once.
The per-link inference cost of \our is thus
\[
C = O(kd + T_{\text{seq}}),
\]
where $k$ is the path length and $d$ the embedding dimension, making it independent of the graph size $n$.
$T_{\text{seq}}$ denotes the cost of a sequence model (e.g., LSTM, Transformer) applied to the node embeddings along that path.

Importantly, since both the GNN embeddings and shortest paths can be precomputed, $B$ is incurred only once. Given that $k$ is typically small in practice, $C$ remains low even at scale. Unlike subgraph-based methods such as SEAL, which require per-link GNN inference over extracted neighborhoods, \our performs lightweight sequence modeling on compact paths, supporting efficient batched inference. Additionally, the framework is flexible: $T_{\text{sp}}$ can be cached or approximated~\cite{sommer2014shortest}, and $T_{\text{seq}}$ depends on the chosen architecture and path length.

We summarize the complexities in the following table:



\begin{table}[h]
\centering
\begin{tabular}{lcc}
\toprule
\textbf{Method} & \textbf{B} & \textbf{C} \\
\midrule
GAE      & $n\Delta d + nd^2$ & $d^2$ \\
Neo-GNN  & $n\Delta d + nd^2 + n\Delta ^l$ & $\Delta^l + d^2$ \\
BUDDY    & $n\Delta d + nh$ & $h + d^2$ \\
SEAL     & $0$ & $\Delta^{(l+1)}d + \Delta^l d^2$ \\
NCN      & $n\Delta d + nd^2$ & $\Delta d + d^2$ \\
NCNC     & $n\Delta d + nd^2$ & $\Delta^2d + \Delta d^2$ \\
PEG      & $n\Delta d + nd^2 + nD^2$ & $d^2$ \\
\textbf{SP4LP} & $n\Delta d + nd^2 + T_{\text{sp}}$ & $kd + T_{\text{seq}}$ \\
\bottomrule
\end{tabular}
\caption{Comparison of precomputation ($B$) and per-link ($C$) complexities across methods.}
\label{tab:complexity}
\end{table}

All methods conform to the form $\mathcal{O}(B + C \cdot t)$, with a one-time graph-level computation and a per-target-link cost. In the table, $h$ is the cost of the hash function used in BUDDY, $l$ denotes the radius of local neighborhoods in Neo-GNN and SEAL, and $D$ denotes the number of Laplacian eigenvectors used for spectral positional encoding in PEG. GAE, NCN, and BUDDY are efficient but limited in expressiveness. NCNC enhances NCN via soft neighbor completion at a slightly higher cost. PEG incorporates spectral encoding ($\mathcal{O}(nD^2)$), while SEAL is the most computationally intensive due to subgraph extraction and per-link GNN processing. SP4LP, in contrast, strikes a balance between efficiency and expressiveness by combining GNN-based embeddings with sequence modeling over shortest paths, whose cost can be mitigated via approximation techniques~\cite{sommer2014shortest}, ensuring scalability for large graphs.

\clearpage

\end{document}